\documentclass{article}
\pdfoutput=1

\usepackage[final,nonatbib]{neurips_2024}

\usepackage[utf8]{inputenc} 
\usepackage[T1]{fontenc}    
\usepackage{hyperref}       
\usepackage{url}            
\usepackage{booktabs}       
\usepackage{amsfonts}       
\usepackage{nicefrac}       
\usepackage{microtype}      
\usepackage{xcolor}         

\usepackage{graphicx}
\usepackage{multirow}
\usepackage{enumitem}
\usepackage{amsmath}
\usepackage{cleveref}
\newcommand\textblue{\textcolor[RGB]{3, 111, 193}}

\usepackage{hyperref}
\hypersetup{
    colorlinks=true,
    linkcolor=red,
    urlcolor=cyan,
    citecolor=cyan,
}
\usepackage{titletoc}

\title{Not All Diffusion Model Activations\\ Have Been Evaluated as Discriminative Features}

\author{\parbox{14cm}{
    \centering
    \large{
        Benyuan Meng$^{1,2}$ \quad
        Qianqian Xu$^{3,4}$\thanks{Corresponding authors.} \quad  
        Zitai Wang$^{3}$
    }
    \\ 
    \large{
        Xiaochun Cao$^{5}$ \quad
        Qingming Huang$^{6,3,7*}$  
    }
    \\
    \normalsize{
        \textnormal{$^1$Institute of Information Engineering, CAS}\\
        \textnormal{$^2$School of Cyber Security, University of Chinese Academy of Sciences}\\
        \textnormal{$^3$Key Lab. of Intelligent Information Processing, Institute of Computing Technology, CAS}\\
        \textnormal{$^4$Peng Cheng Laboratory}\\
        \textnormal{$^5$School of Cyber Science and Tech., Shenzhen Campus of Sun Yat-sen University}\\
        \textnormal{$^6$School of Computer Science and Tech., University of Chinese Academy of Sciences}\\
        \textnormal{$^7$Key Laboratory of Big Data Mining and Knowledge Management, CAS}
    }
    \\
    \texttt{mengbenyuan@iie.ac.cn}\quad
    \texttt{\{xuqianqian,wangzitai\}@ict.ac.cn}\\  
    \texttt{caoxiaochun@mail.sysu.edu.cn}\quad  
    \texttt{qmhuang@ucas.ac.cn}
}}

\begin{document}

\maketitle

\begin{abstract}
  Diffusion models are initially designed for image generation.
  Recent research shows that the internal signals within their backbones, named activations, can also serve as dense features for various discriminative tasks such as semantic segmentation.
  Given numerous activations, selecting a small yet effective subset poses a fundamental problem.
  To this end, the early study of this field performs a large-scale quantitative comparison of the discriminative ability of the activations.
  However, we find that many potential activations have not been evaluated, such as the queries and keys used to compute attention scores.
  Moreover, recent advancements in diffusion architectures bring many new activations, such as those within embedded ViT modules.
  Both combined, activation selection remains unresolved but overlooked.
  To tackle this issue, this paper takes a further step with a much broader range of activations evaluated.
  Considering the significant increase in activations, a full-scale quantitative comparison is no longer operational.
  Instead, we seek to understand the properties of these activations, such that the activations that are clearly inferior can be filtered out in advance via simple qualitative evaluation.
  After careful analysis, we discover three properties universal among diffusion models, enabling this study to go beyond specific models.
  On top of this, we present effective feature selection solutions for several popular diffusion models.
  Finally, the experiments across multiple discriminative tasks validate the superiority of our method over the SOTA competitors.
  Our code is available at \href{https://github.com/Darkbblue/generic-diffusion-feature}{this url}.
\end{abstract}

\section{Introduction}
\label{sec:intro}

Diffusion models~\cite{ho2020denoising, dhariwal2021guideddiffusion, rombach2022high, podell2023sdxl} are powerful generative models that progressively reconstruct images from Gaussian noises through a series of denoising steps.
Typically, a U-Net~\cite{ronneberger2015unet} is trained as the noise predictor backbone to perform denoising.
Recently, the impressive generative capability inspires the application to discriminative tasks such as semantic segmentation~\cite{xu2023open, zhao2023unleashing} or semantic correspondence~\cite{luo2023dhf, zhang2023tale}.
In this direction, \textit{diffusion feature} is one simple yet effective approach, where the intermediate signals, named activations, are extracted from the pre-trained diffusion U-Net as dense features~\cite{baranchuk2021label, zhao2023unleashing, zhang2023tale, luo2023dhf, yang2023diffusion, DBLP:conf/cvpr/DuttMM24, frick2024diffseg}.

The complex architecture of the diffusion U-Net provides many activations that can serve as features.
However, these activations inherently vary in quality, inducing significant performance gaps on discrimination.
Hence, selecting a small yet effective subset from these activations has become a fundamental problem.
In the early stage, Baranchuk \textit{et. al}~\cite{baranchuk2021label} perform a large-scale quantitative comparison among activations within Guided Diffusion~\cite{dhariwal2021guideddiffusion}.
Later, the activations they select are followed by most studies in this field, pursuing other improvements \cite{luo2023dhf, zhang2023tale, zhao2023unleashing, xu2023open, zhou2023hyperspectral, yang2023diffusion, kondapaneni2023text, li2023sd4match}. 

\begin{figure}[t]
  \centering
  \includegraphics[width=.95\linewidth]{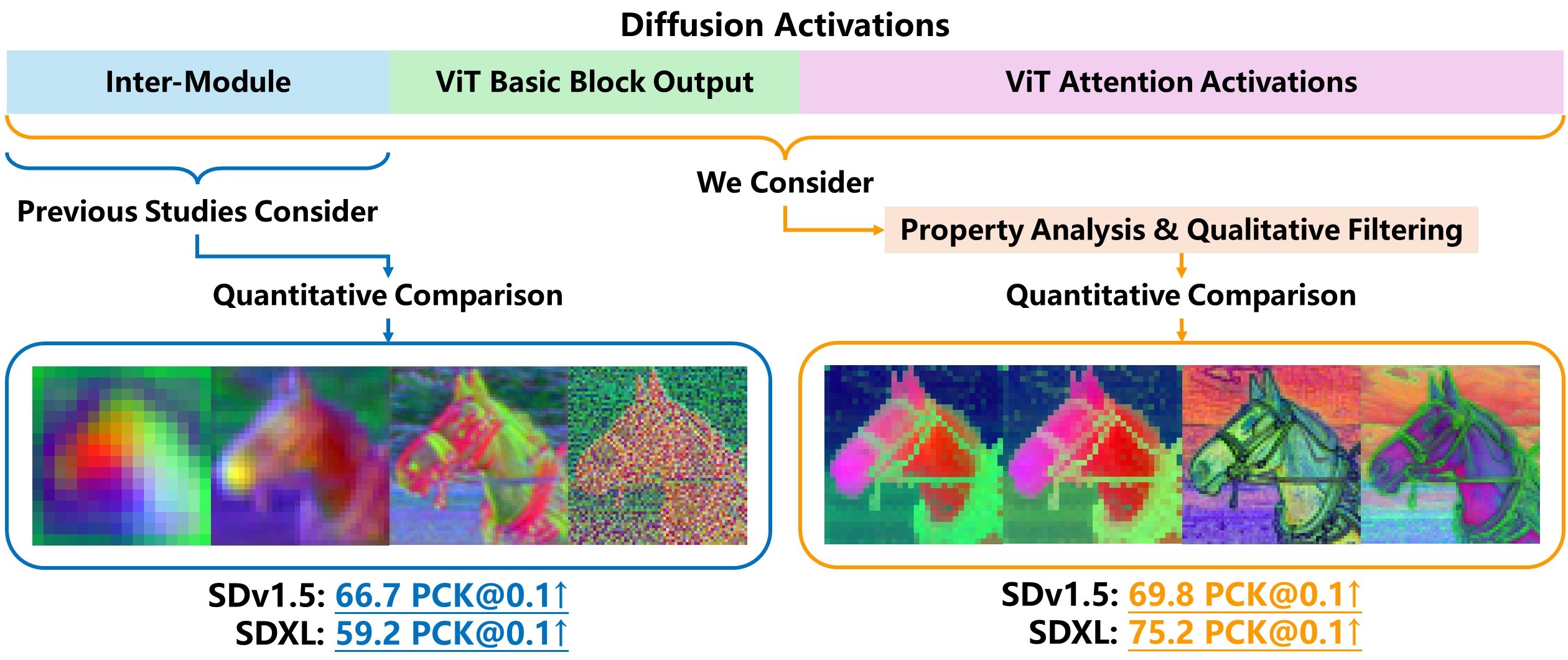}
  \caption{Prior arts only consider a small fraction of potential activations in diffusion models. As a result, more advanced diffusion architecture fails to achieve better performance ({\color[RGB]{12, 113, 189}SDXL \textit{v.s.} SDv1.5}). In contrast, we consider a broader range of candidate activations. To facilitate the quantitative comparison, we first make a comprehensive and generalizable analysis to qualitatively filter out many candidates in advance. On top of this, our method achieves superior performance ({\color[RGB]{247, 164, 31}75.2 PCK@0.1}).
  }
  \label{fig:intro}
\end{figure}

However, we find that this fundamental issue is far from solved. On one hand, Baranchuk \textit{et. al}~\cite{baranchuk2021label} only consider activations between neighboring modules that comprise the main residuals.
This means that many potential activations are excluded from the candidate pool, such as the queries and keys in the self-attention blocks. 
Moreover, recent developments of diffusion architecture, such as cross-attention~\cite{rombach2022high} or embedded deep vision transformers (ViTs)~\cite{podell2023sdxl}, have introduced additional types of activations.
Hence, as shown in upper \Cref{fig:intro}, only a small fraction of potential activations in modern diffusion models have been evaluated for their discriminative ability, which might hinder future work in this direction.
For example, lower \Cref{fig:intro} shows that Stable Diffusion XL (SDXL)~\cite{podell2023sdxl}, which is more advanced than Stable Diffusion v1.5 (SDv1.5)~\cite{rombach2022high}, fails to achieve better performance with the feature selection solution proposed in \cite{baranchuk2021label}. 

In this paper, we revisit the fundamental problem of feature selection, considering a more comprehensive candidate pool of activations.
Due to its large volume, a full-scale quantitative comparison is no longer operational, urging us to modify the previous research methodology in \cite{baranchuk2021label}.
As illustrated in \Cref{fig:intro}, instead of a direct quantitative comparison, we first explore the properties of diffusion U-Nets.
These properties allow us to qualitatively and efficiently filter out many activations that are highly likely to be sub-optimal, shrinking the candidate pool for the quantitative comparison.
More importantly, we find these properties are universal among diffusion models, making it possible to generalize our findings to more models beyond those covered in this paper.

Specifically, the properties we find, which are distinct to existing knowledge of model properties \cite{amir2021deep, ronneberger2015unet}, exactly correspond to three top-to-bottom levels of the diffusion U-Net:
(i) \textbf{Diffusion noises} at the macro level: the diffusion process induces a new type of noise on both low- and high-frequency signals.
(ii) \textbf{In-resolution granularity changes} at the in-resolution level: the changes in information granularity are not only across but also within resolutions.
(iii) \textbf{Locality without positional embeddings} at the sub-module level: the embedded ViTs in diffusion U-Nets present a new type of local information different from that induced by the conventional positional embeddings~\cite{vaswani2017attention}.
Based on these insights, we develop effective feature selection solutions for several popular diffusion models. Finally, the experiments on three discriminative tasks, including semantic correspondence, semantic segmentation, and label-scarce segmentation, validate the superiority of our solutions over the SOTA methods. 

In summary, the contributions of this work are three-fold:
\begin{itemize}[leftmargin=10pt]
    \item \textbf{Revisit of a fundamental problem}: To the best of our knowledge, we are the first to point out that the fundamental issue of feature selection remains unresolved in the realm of diffusion feature.
    \item \textbf{Generic insights}: The properties we find are universal among different diffusion U-Nets, which can provide valuable insights for future work.
    \item \textbf{Extensive validation}: Extensive experiments on three discriminative tasks validate the effectiveness of the feature selection solutions induced by our insights.
\end{itemize}

\section{Related Work}
\label{sec:related}

\noindent \textbf{Diffusion models for discrimination.}
We discuss three mainstream ways to use diffusion models for discrimination.
(i) Diffusion classifier~\cite{li2023your, chen2024your} utilizes Bayes' theorem to transform a pre-trained diffusion model into an image classifier.
This method enjoys a theoretical guarantee and does not need additional training. However, it is limited to image-level tasks.
(ii) The second way is to model discriminative tasks as image-to-image generation tasks with diffusion models~\cite{ji2023ddp, ke2023repurposing}. This method is suitable for various dense vision tasks but requires heavy training.
(iii) Diffusion feature~\cite{baranchuk2021label, zhao2023unleashing, zhang2023tale, tang2023emergent, luo2023dhf}, the focus of this paper, follows the traditional practice of feature extraction to pursue the balance between wider applicability and less training. 
This makes it adaptable for different tasks and alleviates the training needs for the diffusion model.
Only small downstream models may require training.

The diffusion feature approach has seen various improvements.
Some techniques toggle the input settings of diffusion models, such as seeking better timesteps~\cite{zhou2023hyperspectral, luo2023dhf}.
The others add trainable parameters outside the diffusion model to refine the outputs or provide an efficient fine-tuning alternative~\cite{zhao2023unleashing, li2023sd4match, wan2023harnessing}.
Additionally, some studies explore completely training-free methods that utilize spatial attention information~\cite{xiao2023text, tian2023diffuse}, and some attempts focus on text-free diffusion models~\cite{DBLP:journals/corr/abs-2307-08702, DBLP:journals/corr/abs-2311-17921}.

Despite the progress, previous methods only consider activations between neighboring modules, leaving many potential candidates unevaluated.
Our work shows that such an overlook can hinder model performance.
By introducing a more comprehensive feature selection solution, our method could generically enhance both existing and future diffusion feature approaches.

\noindent \textbf{Analysis on model properties.} The inner properties of neural networks have consistently received much attention \cite{carter2019activation, olah2017feature}. 
For transformers, Geva \textit{et al.}~\cite{Geva21ffn} show that the feed-forward layers act as key-value memories and are interpretable for humans.
Amir \textit{et al.}~\cite{amir2021deep} extend this insight to vision transformers and put it into practical applications such as image classification, while Vilas \textit{et, al.}~\cite{Vilas23Analyzing} try to make more detailed interpretation of ViT activations.

The methodology of these studies provides valuable guidance for our research.
However, since diffusion models are trained for generation, our study relies more on qualitative analysis and feature visualization compared to previous work.

\section{Preliminaries: Architecture of Diffusion U-Nets}
\label{sec:pre}

\begin{figure}[t]
  \centering
  \includegraphics[width=\linewidth]{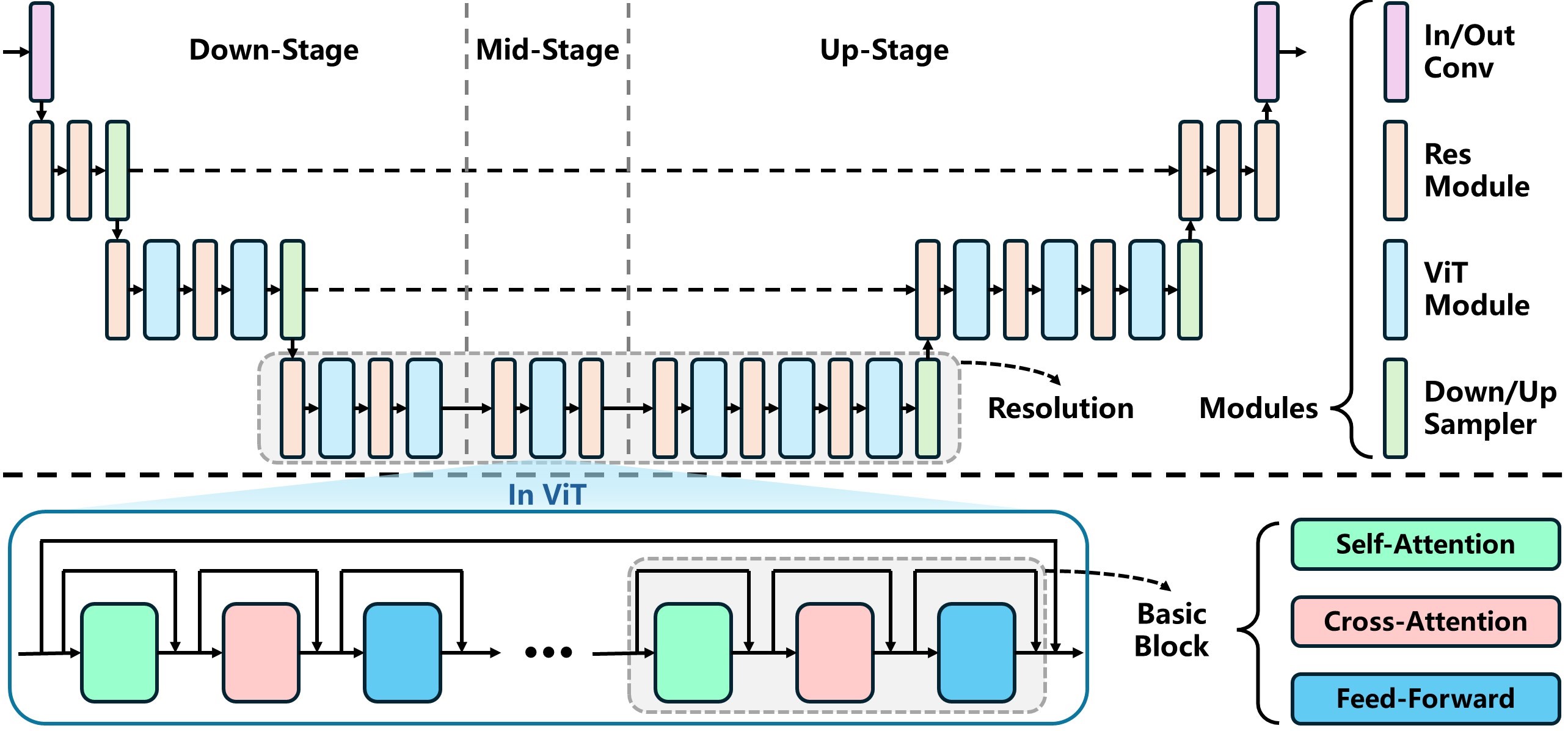}
  \caption{U-Net architecture (upper) and the ViT module (lower), taking SDXL as an example.}
  \label{fig:unet}
\end{figure}

Diffusion models typically involve a forward pass and a reverse pass~\cite{ho2020denoising}. 
During the forward pass, noise is gradually added to a clean image $\boldsymbol{x}_0 \in \mathbb{R}^{3\times w \times h}$ until the image resembles Gaussian noise, where $w$ and $h$ denote the width and height, respectively. This process can be denoted by $\boldsymbol{x}_t \sim q(\boldsymbol{x}_t|\boldsymbol{x}_0)$, where $q$ represents the noise posterior and $t$ is the timestep.
In the reverse pass, an end-to-end neural network $\epsilon_\theta$, as parameterized by $\theta$, learns to predict the noises and thus reconstruct the image.
One such denoising step can be denoted as $\epsilon = \epsilon_\theta(\boldsymbol{x}_t, t, \boldsymbol{c})$,
where $\epsilon$ is the predicted noise, and $\boldsymbol{c}$ is the condition that describes the expected image content.
Although there are alternative formulations for diffusion models~\cite{song2020score, lipman2022flow, su2022dual}, they all rely on this neural network backbone $\epsilon_\theta$.
This study focuses on this backbone, typically implemented as U-Net~\cite{ronneberger2015unet}, rather than other components of diffusion models.
Next, we will detail the architecture of the diffusion U-Net and standardize the \textit{terminology} referring to different parts of the model, as shown in \Cref{fig:unet}.

We start with an overview.
Following the initial U-Net design~\cite{ronneberger2015unet}, the U-Net has three main \textit{stages}: down-stage, mid-stage, and up-stage.
The down-stage reduces the resolution of activations, while the up-stage increases it.
Both stages contain multiple \textit{resolutions}, in each of which the activations share the same resolution.
Furthermore, each resolution includes several \textit{modules}, including \colorbox[RGB]{251, 227, 214}{\textit{ResModule}} (convolutional ResNet~\cite{He16resnet} structures), \colorbox[RGB]{202, 238, 251}{\textit{ViT Module}}~\cite{Dosovitskiy21vit}, and \colorbox[RGB]{217, 242, 208}{\textit{Downsampler/Upsampler}} (simple convolutional layers).
Previous diffusion feature approaches only consider activations between these adjacent modules, \textit{i.e.}, inter-module activations.

We next dive into the details below the module level.
Among these modules, ResModule and ViTs adopt \textit{residual connection}~\cite{He16resnet}, where an increment activation is added element-wise to the residual activation to refine it.
Specifically, ResModule uses simple convolutional layers to produce increments, whereas ViTs use complex attention mechanisms, which will be further explained next.
Typically, ViT operates as a standalone model followed by a decoder that produces the output predictions for visual tasks~\cite{Dosovitskiy21vit,krizhevsky2009learning, deng2009imagenet,hao2020brief}.
However, in the diffusion U-Net, multiple ViTs are embedded into the network, and their outputs serve as increment activations. 

Furthermore, each ViT module consists of several stacked \textit{basic blocks}.
A basic block typically has a \colorbox[RGB]{153, 255, 204}{\textit{self-attention layer}} to perform attention on the image itself and a \colorbox[RGB]{95, 200, 240}{\textit{feed-forward layer}}, essentially a two-layer MLP~\cite{Geva21ffn}. 
Modern diffusion U-Net introduces an additional \colorbox[RGB]{250, 200, 200}{\textit{cross-attention layer}} between the two layers, enabling the fusion of the image and additional textual prompts.
Besides, each layer includes a residual connection, meaning that the increment activation added to the outer residual is also the internal residual.

In a nutshell, the architecture described above provides abundant activations that can serve as dense features. Given the massive activations, it is no longer operational to perform a full-scale quantitative comparison.
Hence, we next make a comprehensive analysis of model properties to better understand these activations, which can help the qualitative filtering.

\section{Distinct Properties of Diffusion U-Nets}
\label{sec:mech}
\begin{figure}[t]
  \centering
  \includegraphics[width=\linewidth]{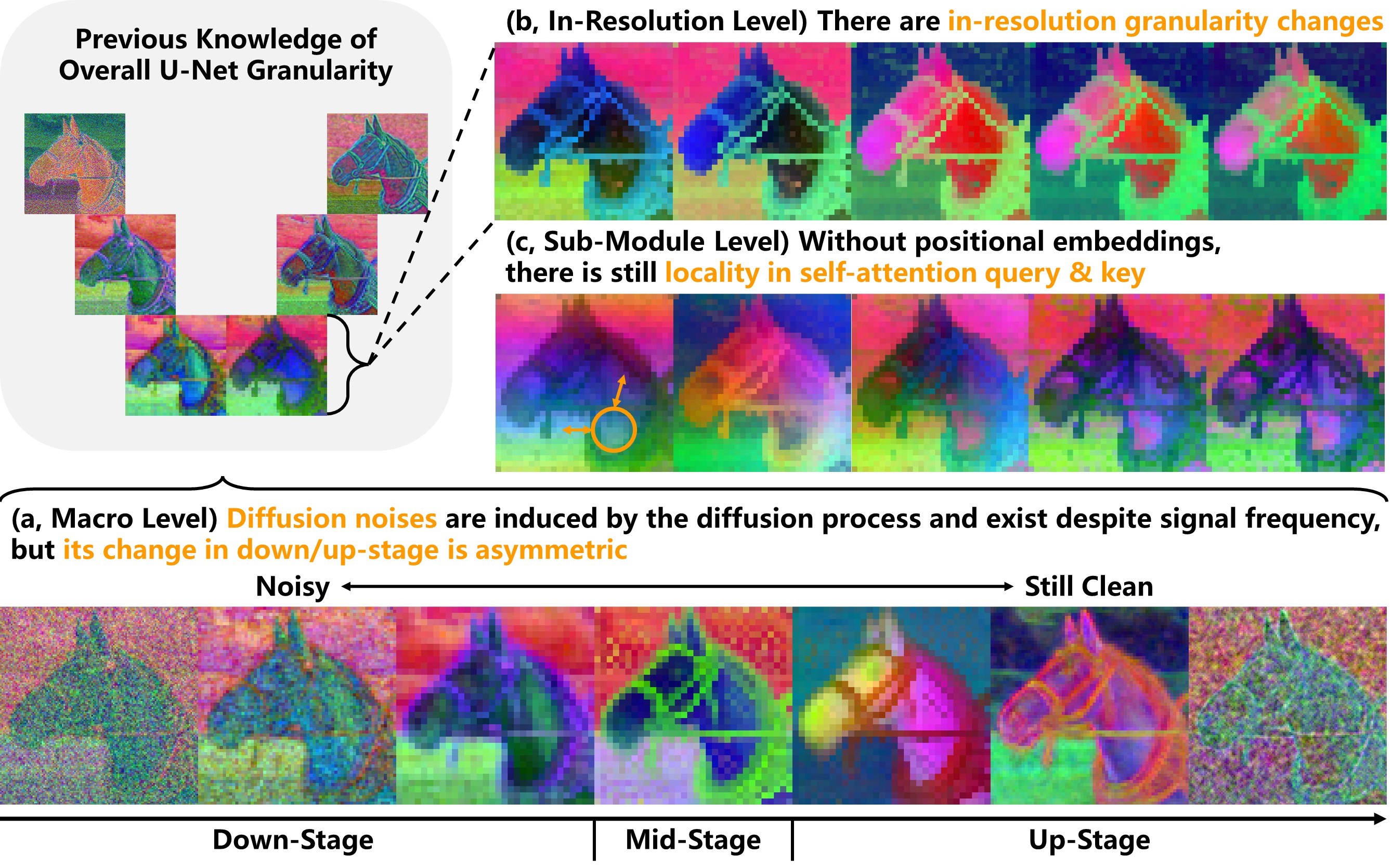}
  \caption{We highlight three properties of diffusion U-Nets that \textbf{are distinct from existing knowledge about other models}:
  (a) Asymmetric diffusion noises.
  (b) In-resolution granularity changes.
  (c) Locality without positional embeddings: pixels within the {\color[RGB]{243, 150, 14}orange circle} resemble nearby background pixels more than distant pixels on the horse's neck that are semantically closer.}
  \label{fig:mech-large}
\end{figure}

The diffusion U-Net has many interesting properties, but now we only focus on those distinct from the knowledge of traditional U-Net~\cite{ronneberger2015unet} or ViTs~\cite{vaswani2017attention}.
As shown in \Cref{fig:mech-large}, this section highlights three noticeable properties, each of which corresponds to a different level of the diffusion U-Net architecture described in \Cref{sec:pre}.
Notably, \textbf{these properties can be universally observed in different samples and diffusion models, though all visualization in the main content is conducted on the same image and SDXL model for consistency}.
Additional visualization is provided in \Cref{sec:app-visualization}. 
Besides, the omitted properties are available in \Cref{sec:app-behavior}.

\subsection{Asymmetric Diffusion Noises}

The first property is at the macro level and closely related to the overall diffusion process.
It is common that high-frequency signals are typically noisy \cite{carter2019activation, olah2017feature}.
This phenomenon can also be observed in diffusion U-Nets, especially within the increment activations of residual connection.
However, this does not mean that low-frequency signal activations in diffusion U-Nets are free from noises. 
As shown in \Cref{fig:mech-large}(a), the diffusion process introduces a new type of noise that also impacts low-frequency signals.
This is not surprising since the diffusion process requires the backbone to process noisy inputs and predict noises as outputs.
As a result, activations near the inputs or outputs, regardless of their frequency, also suffer from such noises.
Considering the special cause, we name such noises \textit{diffusion noises}.

How does the influence of diffusion noises spread across diffusion U-Nets?
As shown in \Cref{fig:mech-large}(a), diffusion noises exist throughout the entire down-stage, with a decreasing magnitude.
Remarkably, during the early half of the up-stage, activations are rather clean, with no perceivable diffusion noises. 
Only in the later half do diffusion noises start to resurface.
The existence of this asymmetric behavior can also be indirectly supported by the ablation curves in many fellow studies such as \cite{baranchuk2021label, DBLP:journals/corr/abs-2307-08702, DBLP:journals/corr/abs-2311-17921}.
Furthermore, as shown in \Cref{sec:app-visualization}, such an asymmetric pattern is consistent across activations in different diffusion models.

Similar to common high-frequency noises, diffusion noises can degenerate feature quality.
Hence, this property can serve as a criterion for identifying and filtering out sub-optimal activations, which we will delve into in \Cref{sec:method}. 

\subsection{In-Resolution Granularity Changes}
The second property is at the in-resolution level and closely related to recent advances in diffusion architectures. 
Specifically, the design of U-Net follows the idea of resolution hierarchy~\cite{ronneberger2015unet}.
Consequently, the overall architecture displays a fine-coarse-fine granularity trend, looking like the alphabet ``U''.
Traditional U-Nets implement this architecture with a relatively large number of resolutions, while each resolution is typically small, equipped with two or three simple convolutional layers.
Hence, the understanding of traditional U-Net focuses on the granularity changes across resolutions, implicitly assuming that the change within a single resolution is negligible.

However, diffusion U-Nets become much ``fatter''.
In other words, modern diffusion U-Nets typically have much fewer resolutions, but each resolution is significantly enlarged. For example, SDv1.5 only has four resolutions \cite{rombach2022high}, and SDXL further decreases this number to three \cite{podell2023sdxl}, as shown in \Cref{fig:unet}. Meanwhile, each resolution can produce much more activations, primarily thanks to the embedded ViT modules. 
This architectural evolution makes the granularity change within a single resolution more significant, as depicted in \Cref{fig:mech-large}(b).

Different granularity carries varied information and quality, resulting in different discriminative performance on downstream tasks~\cite{baranchuk2021label, luo2023dhf}.
Hence, this discovery of \textit{in-resolution granularity changes} highlights the necessity to evaluate more activations, especially those within the embedded ViTs, as they are of different granularity.

\subsection{Locality without Positional Embeddings}

For the third distinct property, we delve into the sub-module level, \textit{i.e.}, the blocks in embedded ViT modules.
Positional embeddings, which are widely used in language transformers~\cite{vaswani2017attention} and ViTs~\cite{Dosovitskiy21vit}, aim to provide spatial information for each input token. Consequently, the activations of traditional ViTs display strong positional information, where the latent pixel resembles nearby pixels more than those that are semantically similar but far away~\cite{amir2021deep}. This phenomenon is significant for the layers close to the inputs. When the layer goes deeper, the tokens are refined with semantic information, making the activations display less positional information. However, only in the last few layers, such positional information becomes negligible. 

In contrast, ViT modules in diffusion U-Nets do not use positional embeddings~\cite{rombach2022high, podell2023sdxl}, perhaps because the other U-Net components have provided sufficient spatial cues.
This change results in distinct properties of the activations. 
On one hand, positional information is negligible for most activations despite how near they are to the inputs.
For example, even in the first basic block, cross-attention query activations contain no perceivable positional information.
On the other hand, the queries and keys of self-attention still display non-negligible positional information, {\color[RGB]{243, 150, 14} marked with orange circles} in \Cref{fig:mech-large}(c).
Specifically, the latent pixel on the horse's neck is a light blue color, similar to the pixels to its left that actually represent the background.
In contrast, the pixels above the circle are in purple color, though they also represent the horse's neck.
Such comparison shows that a latent pixel is more similar to other pixels that are spatially near it than those semantically closer to it.

We name this phenomenon \textit{locality} since it has a different mechanism from that induced by positional embeddings.
As pointed out in \cite{Dosovitskiy21vit}, self-attention allows ViT to integrate global and local information even in the shallow layers, and the attention scope enlarges \textit{w.r.t.} layer depth.
Even without positional embeddings, self-attention activations are generally consistent with this insight, leading to the existence of locality.
Nevertheless, the magnitude has indeed greatly reduced, compared to the visualization of conventional ViT activations in ~\cite{amir2021deep}.
As shown in \Cref{fig:mech-large}(c), in shallow activations, locality exists but is inherently weaker, as much semantic information is still reserved.
In addition, locality quickly diminishes as the layer goes deeper.

Since positional information can degrade the quality of activations~\cite{amir2021deep}, its absence has the potential to enhance the activations in ViT modules.
Moreover, locality can play a special role in activation filtering, as presented in \Cref{sec:method}.

\subsection{Universality of Three Properties}
Although the visualization in \Cref{fig:mech-large} is conducted only on SDXL, the scope of these properties is not limited to the specific architecture. Further supporting evidence is available in \Cref{sec:app-visualization}.
\begin{enumerate}[label=(\roman*), leftmargin=25pt]
    \item Diffusion noises directly arise from the diffusion process. Hence, it is promising to extend this property to other diffusion backbones.
    \item In-resolution granularity changes come from the ``fatness'' of U-Nets, making it potentially applicable to more traditional U-Nets.
    \item Locality originates from the self-attention mechanism in ViT architectures, so it is broadly applicable to standalone or embedded ViTs where positional embeddings are absent.
\end{enumerate}

\section{Enhanced Feature Selection from Diffusion U-Nets}
\label{sec:method}

So far, we have had a more comprehensive understating of the properties of diffusion U-Nets.
All these properties, especially in-resolution granularity changes, encourage us to reconsider the feature selection solution, with a special emphasis on the activations in ViT modules.
With these properties, we are also able to filter out many low-quality activations qualitatively, followed by a thus simplified quantitative comparison.

\begin{figure}[t]
  \centering
  \includegraphics[width=\linewidth]{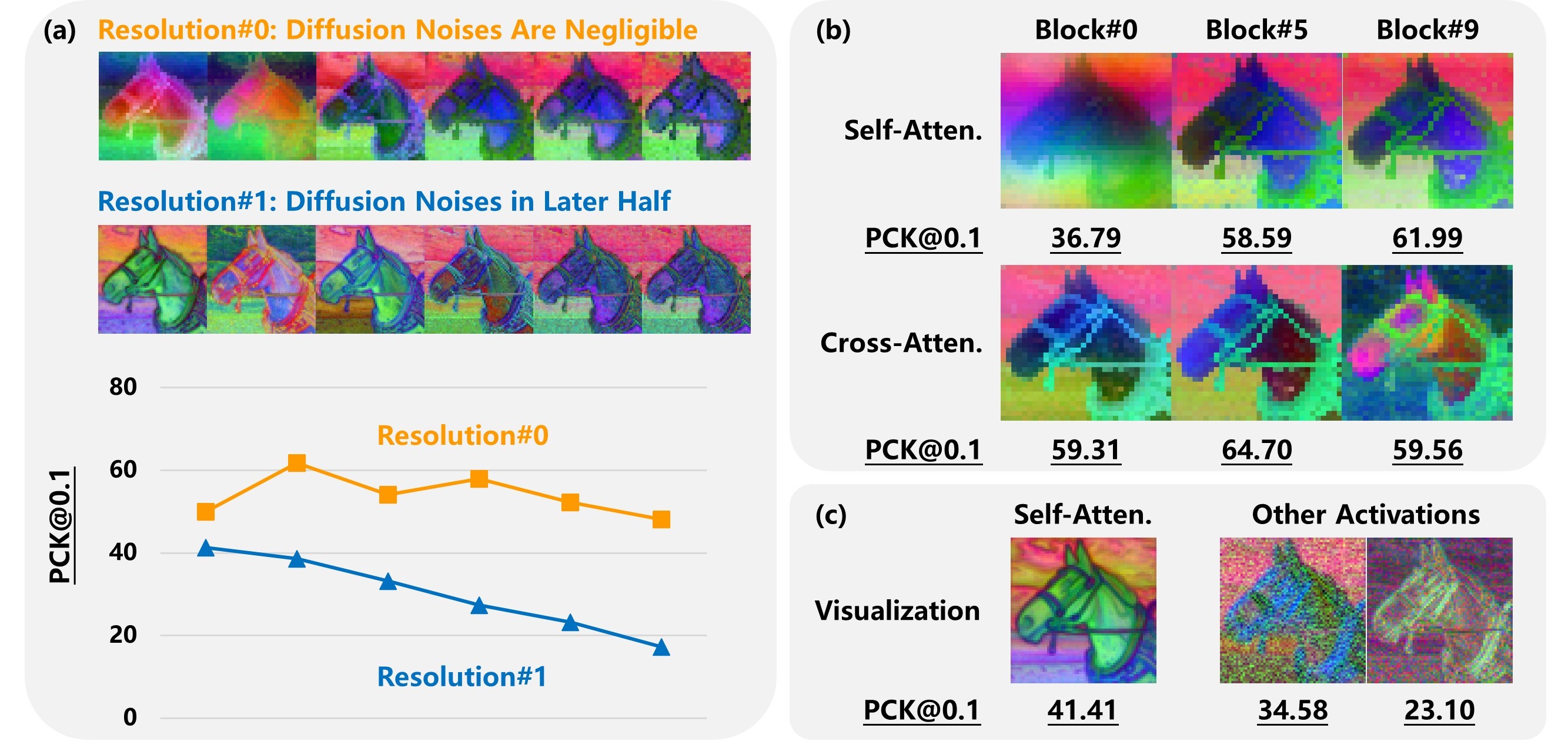}
  \caption{(a) Diffusion noises result in a significant performance degeneration ({\color[RGB]{4, 112, 186} Resolution\#1}).
  (b) Locality degrades the quality of self-attention activations (Block\#0 and Block\#5).
  (c) Locality in self-attention activations can suppress diffusion noises, leading to better quality than noisy activations (41.41 \textit{v.s.} 34.58).
  All $\text{PCK@0.1}_{\text{img}}(\uparrow)$ results are evaluated on the semantic correspondence task.}
  \label{fig:evidence}
\end{figure}

\subsection{Qualitative Filtering}
\label{sec:method-1}
\textbf{Avoiding Diffusion Noises.}
As shown in \Cref{fig:evidence}(a), diffusion noises tend to degrade the quality of activations.
Hence, it is natural to filter out the activations severely affected by diffusion noises from the candidate pool. 
Specifically, according to the asymmetric trend of diffusion noises, \textbf{we only consider activations in the early half of the up-stage}, which are rather clean.
This approach will significantly reduce the number of candidate activations and simplify the quantitative comparison.

\textbf{Avoiding Self-Attention Locality.}
The locality in self-attention modules is another important factor that can degrade the activations.
The empirical evidence in \Cref{fig:evidence}(b) demonstrates that these activations are generally inferior to the others, such as those from cross-attention layers or the outputs of ViT basic blocks.
Consequently, it is rather safe to \textbf{filter out most activations in self-attention modules} from our candidate pool.

\textbf{Using Locality to Suppress Diffusion Noises.}
So far, the activations in the candidate pool are clean and free from locality.
However, all these activations are low-resolution ones since high-resolution activations are generally noisy and thus filtered out.
This is unfavorable since some detailed information might only exist in high-resolution activations.
To address this issue, we exploit a side effect of self-attention locality.
Specifically, as indicated in \Cref{fig:evidence}(c), locality can help suppress diffusion noises via its focus on spatial structures.
Although locality is sub-optimal, it is still superior to severe diffusion noises. In view of this, the candidate pool \textbf{reserves self-attention activations extracted from the later half of the up-stage}. 

We have filtered out many candidate activations based on the distinct properties of diffusion U-Nets. Additionally, all increment activations in residual connection can be further filtered out since they introduce high-frequency noises \cite{carter2019activation, olah2017feature}. 
After such qualitative filtering, a small but high-quality candidate pool is available. Taking SDXL as an example, the number of candidates decreases from 279 to 63, \textit{i.e.}, a 78\% reduction.
Next, we explain how to conduct this quantitative comparison briefly.

\subsection{Quantitative Comparison}
The quantitative comparison follows the protocol of \cite{baranchuk2021label}. Specifically, given an input image $\boldsymbol{x}_0$, we first perform the forward pass with a pre-defined timestep $t$ to generate the noisy image $\boldsymbol{x}_t$. Then, the U-Net backbone conducts one denoising step. Instead of collecting the model output $\epsilon$, we gather U-Net activations and consolidate them to the candidate pool, as described in \Cref{sec:method-1}.
Afterward, each activation is individually fed to a downstream model to evaluate its discriminative potential, with the details of the model described in \Cref{sec:app-detail}.
Such comparison is made fair by using the same dataset for all activations.
Finally, we can obtain the capability ranking of activations from each resolution, according to which features can be selected wisely.
Notably, we conduct such comparisons on multiple datasets to guarantee generalizability, since the best activations may differ among different scenes~\cite{baranchuk2021label}.
Thanks to qualitative filtering, \textbf{the time cost for each (model, dataset) pair has been reduced by more than one week}, equipped with Nvidia(R) RTX 3090 GPUs.

Given the capability ranking, feature selection is in fact flexible, as it is possible to combine multiple activations and specifically tune the choice for a task.
For practicality, we provide off-the-shelf feature combinations for SDv1.5 and SDXL that are likely to generically perform well in \Cref{sec:app-selection}, according to the results detailed in \Cref{sec:app-exp}.
\textbf{Each of them consists of four activations mainly from ViT modules rather than inter-module positions}.
Taking SDv1.5 as an example, one of the four selected features is from one self-attention layer in the highest resolution, which utilizes our observation in \Cref{sec:mech}.
\section{Experimental Validation on Multiple Discriminative Tasks}
\label{sec:exp}

\begin{table}[t]
\centering
\caption{Experimental results on the semantic correspondence task. The best results are in \textcolor{orange}{\textbf{bold}} font and the runner-up is \textblue{\underline{underlined}}.}
\label{tab:exp-corres}
\begin{tabular}{@{}llcc@{}}
\toprule
Category              & Method        & $\text{PCK@0.1}_{\text{img}}\uparrow$        & $\text{PCK@0.1}_{\text{bbox}}\uparrow$       \\ \midrule
\multirow{4}{*}{SOTA} & DINO          & 51.68          & 41.04          \\
                      & DHPF          & 55.28          & 42.63          \\
                      & DIFT          & -              & 52.90          \\
                      & DHF           & 72.56          & 64.61          \\ \midrule
\multirow{2}{*}{Baseline}&Legacy-v1.5 & 75.14          & 66.73          \\
                      & Legacy-XL     & 66.00          & 59.16          \\ \midrule
\multirow{3}{*}{Ours} & Ours-v1.5     & 77.78          & 69.83          \\
                      & Ours-XL       & \textblue{\underline{81.72}}    & \textblue{\underline{75.18}}    \\
                      & Ours-XL-t    & \textcolor{orange}{\textbf{83.90}} & \textcolor{orange}{\textbf{76.86}} \\ \bottomrule
\end{tabular}
\end{table}

To validate the effectiveness of our feature selection solution, the experiments are conducted on three popular discriminative tasks: semantic correspondence, semantic segmentation, and label-scarce segmentation. The SOTA methods for each task are selected as competitors.
Besides, we also compare with two baselines that select the conventional inter-module activations as features~\cite{baranchuk2021label}, named \textit{Legacy-v1.5} and \textit{Legacy-XL}. For our method, we provide the following implementations:
\begin{itemize}[leftmargin=25pt]
    \item \textit{Ours-v1.5 \& Ours-XL}: Features extracted from SDv1.5 and SDXL, respectively.
    \item \textit{Ours-XL-t}: For fairness, we further enhance SDXL features with some additional techniques that are also adopted by SOTAs. The techniques we select, \textit{i.e.}, attention maps~\cite{zhao2023unleashing} and feature amalgamation~\cite{baranchuk2021label, luo2023dhf, zhou2023hyperspectral}, are relatively simple and lightweight.
\end{itemize}
More experimental details are in \Cref{sec:app-detail}, such as task information, evaluation metrics, and implementation details.
Besides, \Cref{sec:app-exp} presents additional results not covered here.

\subsection{Empirical Results on Semantic Correspondence}
We present the experimental results for semantic correspondence in \Cref{tab:exp-corres}. From the results, we have the following observations:
\begin{enumerate}[label=(\roman*), leftmargin=25pt]
    \item \textit{Ours-v1.5} outperforms \textit{Legacy-v1.5} (77.78 \textit{v.s.} 75.14 on $\text{PCK@0.1}_{\text{img}}$).
    The main reason is that \textit{Legacy-v1.5} fails to effectively handle the diffusion noises in high-resolution activations.
    Previous approaches either reserve the noisy activations and thus suffer from performance degradation~\cite{baranchuk2021label, xu2023open}, or simply discard high-resolution activations and thus suffer from information loss~\cite{zhao2023unleashing, wan2023harnessing}.
    In contrast, our approach uses self-attention locality to suppress diffusion noises and harvest better high-resolution activations.
    \item \textit{Legacy-XL} is inferior to \textit{Legacy-v1.5}  (66.00 \textit{v.s.} 75.14 on $\text{PCK@0.1}_{\text{img}}$). At first glance, this result is counter-intuitive since SDXL is more advanced than SDv1.5. However, the analysis in \Cref{sec:mech} can unravel the mystery.
    Specifically, since SDXL has more ViT modules, the more valuable activations shift from inter-module positions to these embedded ViTs. Since the baseline does not consider ViT modules, \textit{Legacy-XL} fails to achieve better performance.
    In contrast, \textit{Ours-XL} shows improvement over \textit{Ours-v1.5} (\textblue{\underline{81.72}} \textit{v.s.} 77.78 on $\text{PCK@0.1}_{\text{img}}$).
    This is consistent with the advance in model architecture and again validates our analysis.
    \item \textit{Ours-XL-t} significantly outperforms the SOTA method with a similar amalgamation technique, \textit{i.e.}, DHF~\cite{luo2023dhf} (\textcolor{orange}{\textbf{83.90}} \textit{v.s.} 72.56 on $\text{PCK@0.1}_{\text{img}}$). This performance gain again validates the effectiveness of our method.
\end{enumerate}

\begin{table}[t]
\centering
\caption{Experimental results on semantic segmentation and its altered version with scarce labeled data, evaluated using mIoU$\uparrow$ metric. The best results are in \textcolor{orange}{\textbf{bold}} font and the runner-up is \textblue{\underline{underlined}}.}
\label{tab:exp-segm}
\begin{tabular}{@{}l|lcc|lc@{}}
\toprule
\multirow{2}{*}{Category} & \multirow{2}{*}{Method} & \multicolumn{2}{c|}{Standard Setting} & \multirow{2}{*}{Method} & Label-Scarce Setting \\
                          &              & ADE20K & CityScapes &             & Horse-21            \\ \midrule
\multirow{4}{*}{SOTA}     & MaskCLIP     & 23.70  & -          & SwAVw2      & 54.0 ± 0.9          \\
                          & ODISE        & 29.90  & -          & MAE         & 63.4 ± 1.4          \\
                          & VPD          & 37.63  & 55.06      & DatasetDDPM & 60.8 ± 1.0          \\
                          & Meta Prompts & 40.89  & 71.94      & DDPM        & \textblue{\underline{65.0 ± 0.8}} \\ \midrule
\multirow{2}{*}{Baseline} & Legacy-v1.5  & 40.26  & 64.01      & Legacy-v1.5 & 59.4 ± 1.3          \\
                          & Legacy-XL    & 27.78  & 71.67      & Legacy-XL   & 53.0 ± 0.9          \\ \midrule
\multirow{3}{*}{Ours}     & Ours-v1.5    & 41.07  & 64.10      & Ours-v1.5   & 60.2 ± 0.9          \\
                          & Ours-XL      & \textblue{\underline{43.45}}  & \textblue{\underline{74.47}}      & Ours-XL     & 62.7 ± 0.7          \\
                          & Ours-XL-t   & \textcolor{orange}{\textbf{45.71}}  & \textcolor{orange}{\textbf{75.89}}      & Ours-XL-t  & \textcolor{orange}{\textbf{66.3 ± 0.9}}\\ \bottomrule
\end{tabular}
\end{table}

\subsection{Empirical Results on Semantic Segmentation}
As shown in the left part of \Cref{tab:exp-segm}, \textit{Ours-XL-t} and \textit{Ours-XL} achieve state-of-the-art performance on the semantic segmentation task (\textcolor{orange}{\textbf{45.71}} and \textblue{\underline{43.45}} \textit{v.s.} 40.89 on ADE20K), demonstrating its effectiveness and generalizability.
Furthermore, the most competitive SOTA, Meta Prompts~\cite{wan2023harnessing}, introduces a large number of trainable parameters and uses the diffusion U-Net recurrently, which is rather time-consuming.
In contrast, our method delivers superior results with efficiency maintained.

Unlike the results in semantic correspondence, \textit{Legacy-XL} outperforms \textit{Legacy-v1.5} on CityScapes (71.67 \textit{v.s.} 64.01), and the performance gap between \textit{Legacy-v1.5} and \textit{Ours-v1.5} is narrow (64.01 \textit{v.s.} 64.10).
This is because this task utilizes a relatively large-scale downstream model, which can significantly refine the input features and thus reduce the gap in feature quality.
Nevertheless, \textit{Ours-XL} still achieves a significant improvement over \textit{Legacy-XL} (\textblue{\underline{74.47}} \textit{v.s.} 71.67).

\subsection{Empirical Results on Label-Scarce Segmentation}
One advantage of diffusion features is the applicability to label-scarce scenarios~\cite{baranchuk2021label}.
For validation under such conditions, we experiment on the label-scarce segmentation task, with results presented in the right part of \Cref{tab:exp-segm}.
The observations are generally similar to those on the semantic correspondence task.
For example, \textit{Ours-v1.5} outperforms \textit{Legacy-v1.5} (60.2 \textit{v.s.} 59.4),
\textit{Legacy-XL} is inferior to \textit{Legacy-v1.5} (53.0 \textit{v.s.} 59.4),
and \textit{Ours-XL} is better than \textit{Ours-v1.5} (62.7 \textit{v.s.} 60.2).

Next, we focus on the comparison with the SOTA method, DDPM~\cite{baranchuk2021label}.
Although DDPM is a relatively early study, it outperforms the other competitors and our most implementations.
This result has two reasons.
On one hand, DDPM performs the diffusion process directly in the image space rather than the currently common practice of compressed latent space.
Although inefficient, such an implementation yields better discriminative features.
On the other hand, DDPM utilizes diffusion models specifically trained on each dataset.
In comparison, we use pre-trained general-purposed SDv1.5 or SDXL to be more efficient and thus consistent with the motivation of diffusion feature. 
Hence, it is rather challenging to surpass this SOTA method.
Fortunately, our best implementation, \textit{Ours-XL-t}, achieves this goal with the help of additional lightweight techniques (\textcolor{orange}{\textbf{66.3}} \textit{v.s.} \textblue{\underline{65.0}}).

\section{Conclusion and Future Work}
\label{sec:conclusion}

In this study, we revisit the fundamental problem of feature selection from diffusion U-Nets.
We point out that prior arts only consider a limited range of potential activations.
In contrast, we consider a much wider range of activations as candidates, especially those extracted from the embedded ViT modules.
Given the large volume of the candidate pool, we first analyze the properties of diffusion U-Nets.
The properties we find are universal such that our observations are not limited to the specific diffusion architecture. 
Based on these properties, we qualitatively filter out many activations with low quality, facilitating the following quantitative comparison.
On top of this, concrete feature selection solutions are proposed for two popular diffusion models, \textit{i.e.}, SDv1.5 and SDXL. 
Finally, extensive experiments on three discriminative tasks validate the effectiveness of our method.

However, we are not sure whether our observations can generalize well to recently-developed DiT models~\cite{Peebles23dit} since they have a markedly different architecture from U-Net-based diffusion models.
Thus, analyzing DiT models is a promising topic for future research.

\section*{Acknowledgments}
This work was supported in part by the National Key R\&D Program of China under Grant 2018AAA0102000, in part by National Natural Science Foundation of China: 62236008, U21B2038, U23B2051, 61931008, 62122075 and 62025604, in part by Youth Innovation Promotion Association CAS, in part by the Strategic Priority Research Program of the Chinese Academy of Sciences, Grant No. XDB0680000, in part by the Innovation Funding of ICT, CAS under Grant No.E000000, in part by the China National Postdoctoral Program for Innovative Talents under Grant BX20240384.

\bibliography{neurips_2024}

\begin{thebibliography}{10}

\bibitem{amir2021deep}
Shir Amir, Yossi Gandelsman, Shai Bagon, and Tali Dekel.
\newblock Deep vit features as dense visual descriptors.
\newblock {\em ECCV Workshop on What is Motion For}, 2022.

\bibitem{baranchuk2021label}
Dmitry Baranchuk, Andrey Voynov, Ivan Rubachev, Valentin Khrulkov, and Artem Babenko.
\newblock Label-efficient semantic segmentation with diffusion models.
\newblock In {\em The Tenth International Conference on Learning Representations}, 2022.

\bibitem{caron2020unsupervised}
Mathilde Caron, Ishan Misra, Julien Mairal, Priya Goyal, Piotr Bojanowski, and Armand Joulin.
\newblock Unsupervised learning of visual features by contrasting cluster assignments.
\newblock In {\em Annual Conference on Neural Information Processing Systems}, 2020.

\bibitem{dino}
Mathilde Caron, Hugo Touvron, Ishan Misra, Herv{\'{e}} J{\'{e}}gou, Julien Mairal, Piotr Bojanowski, and Armand Joulin.
\newblock Emerging properties in self-supervised vision transformers.
\newblock In {\em {IEEE/CVF} International Conference on Computer Vision, {ICCV}}, pages 9630--9640, 2021.

\bibitem{carter2019activation}
Shan Carter, Zan Armstrong, Ludwig Schubert, Ian Johnson, and Chris Olah.
\newblock Activation atlas.
\newblock {\em Distill}, 4(3):e15, 2019.

\bibitem{chen2024videocrafter2}
Haoxin Chen, Yong Zhang, Xiaodong Cun, Menghan Xia, Xintao Wang, Chao Weng, and Ying Shan.
\newblock Videocrafter2: Overcoming data limitations for high-quality video diffusion models.
\newblock {\em CoRR}, abs/2401.09047, 2024.

\bibitem{chen2024your}
Huanran Chen, Yinpeng Dong, Shitong Shao, Zhongkai Hao, Xiao Yang, Hang Su, and Jun Zhu.
\newblock Your diffusion model is secretly a certifiably robust classifier.
\newblock {\em CoRR}, abs/2402.02316, 2024.

\bibitem{Cordts16city}
Marius Cordts, Mohamed Omran, Sebastian Ramos, Timo Rehfeld, Markus Enzweiler, Rodrigo Benenson, Uwe Franke, Stefan Roth, and Bernt Schiele.
\newblock The cityscapes dataset for semantic urban scene understanding.
\newblock In {\em {IEEE} Conference on Computer Vision and Pattern Recognition}, pages 3213--3223, 2016.

\bibitem{deng2009imagenet}
Jia Deng, Wei Dong, Richard Socher, Li{-}Jia Li, Kai Li, and Li~Fei{-}Fei.
\newblock Imagenet: {A} large-scale hierarchical image database.
\newblock In {\em {IEEE} Computer Society Conference on Computer Vision and Pattern Recognition}, pages 248--255, 2009.

\bibitem{dhariwal2021guideddiffusion}
Prafulla Dhariwal and Alexander~Quinn Nichol.
\newblock Diffusion models beat gans on image synthesis.
\newblock In {\em Annual Conference on Neural Information Processing Systems}, pages 8780--8794, 2021.

\bibitem{Dong23maskclip}
Zheng Ding, Jieke Wang, and Zhuowen Tu.
\newblock Open-vocabulary universal image segmentation with maskclip.
\newblock In {\em International Conference on Machine Learning, {ICML}}, volume 202 of {\em Proceedings of Machine Learning Research}, pages 8090--8102, 2023.

\bibitem{Dosovitskiy21vit}
Alexey Dosovitskiy, Lucas Beyer, Alexander Kolesnikov, Dirk Weissenborn, Xiaohua Zhai, Thomas Unterthiner, Mostafa Dehghani, Matthias Minderer, Georg Heigold, Sylvain Gelly, Jakob Uszkoreit, and Neil Houlsby.
\newblock An image is worth 16x16 words: Transformers for image recognition at scale.
\newblock In {\em 9th International Conference on Learning Representations}, 2021.

\bibitem{DBLP:conf/cvpr/DuttMM24}
Niladri~Shekhar Dutt, Sanjeev Muralikrishnan, and Niloy~J. Mitra.
\newblock Diffusion 3d features (diff3f) decorating untextured shapes with distilled semantic features.
\newblock In {\em {IEEE/CVF} Conference on Computer Vision and Pattern Recognition}, pages 4494--4504, 2024.

\bibitem{frick2024diffseg}
Raphael~Antonius Frick and Martin Steinebach.
\newblock Diffseg: Towards detecting diffusion-based inpainting attacks using multi-feature segmentation.
\newblock In {\em {IEEE/CVF} Conference on Computer Vision and Pattern Recognition}, pages 3802--3808, 2024.

\bibitem{Geva21ffn}
Mor Geva, Roei Schuster, Jonathan Berant, and Omer Levy.
\newblock Transformer feed-forward layers are key-value memories.
\newblock In {\em Proceedings of the 2021 Conference on Empirical Methods in Natural Language Processing}, pages 5484--5495, 2021.

\bibitem{DBLP:journals/corr/abs-2307-02138}
Rui Gong, Martin Danelljan, Han Sun, Julio~Delgado Mangas, and Luc~Van Gool.
\newblock Prompting diffusion representations for cross-domain semantic segmentation.
\newblock {\em CoRR}, abs/2307.02138, 2023.

\bibitem{han2024aucseg}
Boyu Han, Qianqian Xu, Zhiyong Yang, Shilong Bao, Peisong Wen, Yangbangyan Jiang, and Qingming Huang.
\newblock Aucseg: Auc-oriented pixel-level long-tail semantic segmentation, 2024.

\bibitem{hao2020brief}
Shijie Hao, Yuan Zhou, and Yanrong Guo.
\newblock A brief survey on semantic segmentation with deep learning.
\newblock {\em Neurocomputing}, 406:302--321, 2020.

\bibitem{he2022masked}
Kaiming He, Xinlei Chen, Saining Xie, Yanghao Li, Piotr Doll{\'{a}}r, and Ross~B. Girshick.
\newblock Masked autoencoders are scalable vision learners.
\newblock In {\em {IEEE/CVF} Conference on Computer Vision and Pattern Recognition}, pages 15979--15988, 2022.

\bibitem{He16resnet}
Kaiming He, Xiangyu Zhang, Shaoqing Ren, and Jian Sun.
\newblock Deep residual learning for image recognition.
\newblock In {\em {IEEE} Conference on Computer Vision and Pattern Recognition}, pages 770--778, 2016.

\bibitem{ho2020denoising}
Jonathan Ho, Ajay Jain, and Pieter Abbeel.
\newblock Denoising diffusion probabilistic models.
\newblock In {\em Annual Conference on Neural Information Processing Systems}, 2020.

\bibitem{ji2023ddp}
Yuanfeng Ji, Zhe Chen, Enze Xie, Lanqing Hong, Xihui Liu, Zhaoqiang Liu, Tong Lu, Zhenguo Li, and Ping Luo.
\newblock {DDP:} diffusion model for dense visual prediction.
\newblock In {\em {IEEE/CVF} International Conference on Computer Vision}, pages 21684--21695, 2023.

\bibitem{ke2023repurposing}
Bingxin Ke, Anton Obukhov, Shengyu Huang, Nando Metzger, Rodrigo~Caye Daudt, and Konrad Schindler.
\newblock Repurposing diffusion-based image generators for monocular depth estimation.
\newblock {\em CoRR}, abs/2312.02145, 2023.

\bibitem{kondapaneni2023text}
Neehar Kondapaneni, Markus Marks, Manuel Knott, Rog{\'{e}}rio Guimar{\~{a}}es, and Pietro Perona.
\newblock Text-image alignment for diffusion-based perception.
\newblock {\em CoRR}, abs/2310.00031, 2023.

\bibitem{krizhevsky2009learning}
Alex Krizhevsky, Geoffrey Hinton, et~al.
\newblock Learning multiple layers of features from tiny images.
\newblock 2009.

\bibitem{li2023your}
Alexander~C. Li, Mihir Prabhudesai, Shivam Duggal, Ellis Brown, and Deepak Pathak.
\newblock Your diffusion model is secretly a zero-shot classifier.
\newblock In {\em {IEEE/CVF} International Conference on Computer Vision}, pages 2206--2217, 2023.

\bibitem{li2023sd4match}
Xinghui Li, Jingyi Lu, Kai Han, and Victor Prisacariu.
\newblock Sd4match: Learning to prompt stable diffusion model for semantic matching.
\newblock {\em CoRR}, abs/2310.17569, 2023.

\bibitem{lipman2022flow}
Yaron Lipman, Ricky T.~Q. Chen, Heli Ben{-}Hamu, Maximilian Nickel, and Matthew Le.
\newblock Flow matching for generative modeling.
\newblock In {\em The Eleventh International Conference on Learning Representations}, 2023.

\bibitem{luo2023dhf}
Grace Luo, Lisa Dunlap, Dong~Huk Park, Aleksander Holynski, and Trevor Darrell.
\newblock Diffusion hyperfeatures: Searching through time and space for semantic correspondence.
\newblock In {\em Annual Conference on Neural Information Processing Systems}, 2023.

\bibitem{min2019spair}
Juhong Min, Jongmin Lee, Jean Ponce, and Minsu Cho.
\newblock Spair-71k: {A} large-scale benchmark for semantic correspondence.
\newblock {\em CoRR}, abs/1908.10543, 2019.

\bibitem{min2020learning}
Juhong Min, Jongmin Lee, Jean Ponce, and Minsu Cho.
\newblock Learning to compose hypercolumns for visual correspondence.
\newblock In {\em Computer Vision - {ECCV} 2020 - 16th European Conference}, volume 12360, pages 346--363. Springer, 2020.

\bibitem{DBLP:journals/corr/abs-2307-08702}
Soumik Mukhopadhyay, Matthew Gwilliam, Vatsal Agarwal, Namitha Padmanabhan, Archana Swaminathan, Srinidhi Hegde, Tianyi Zhou, and Abhinav Shrivastava.
\newblock Diffusion models beat gans on image classification.
\newblock {\em CoRR}, abs/2307.08702, 2023.

\bibitem{DBLP:journals/corr/abs-2311-17921}
Soumik Mukhopadhyay, Matthew Gwilliam, Yosuke Yamaguchi, Vatsal Agarwal, Namitha Padmanabhan, Archana Swaminathan, Tianyi Zhou, and Abhinav Shrivastava.
\newblock Do text-free diffusion models learn discriminative visual representations?
\newblock {\em CoRR}, abs/2311.17921, 2023.

\bibitem{olah2017feature}
Chris Olah, Alexander Mordvintsev, and Ludwig Schubert.
\newblock Feature visualization.
\newblock {\em Distill}, 2(11):e7, 2017.

\bibitem{Peebles23dit}
William Peebles and Saining Xie.
\newblock Scalable diffusion models with transformers.
\newblock In {\em {IEEE/CVF} International Conference on Computer Vision}, pages 4172--4182, 2023.

\bibitem{podell2023sdxl}
Dustin Podell, Zion English, Kyle Lacey, Andreas Blattmann, Tim Dockhorn, Jonas M{\"{u}}ller, Joe Penna, and Robin Rombach.
\newblock {SDXL:} improving latent diffusion models for high-resolution image synthesis.
\newblock {\em CoRR}, abs/2307.01952, 2023.

\bibitem{rombach2022high}
Robin Rombach, Andreas Blattmann, Dominik Lorenz, Patrick Esser, and Bj{\"{o}}rn Ommer.
\newblock High-resolution image synthesis with latent diffusion models.
\newblock In {\em {IEEE/CVF} Conference on Computer Vision and Pattern Recognition}, pages 10674--10685, 2022.

\bibitem{ronneberger2015unet}
Olaf Ronneberger, Philipp Fischer, and Thomas Brox.
\newblock U-net: Convolutional networks for biomedical image segmentation.
\newblock In {\em Medical Image Computing and Computer-Assisted Intervention}, volume 9351, pages 234--241. Springer, 2015.

\bibitem{song2020score}
Yang Song, Jascha Sohl{-}Dickstein, Diederik~P. Kingma, Abhishek Kumar, Stefano Ermon, and Ben Poole.
\newblock Score-based generative modeling through stochastic differential equations.
\newblock In {\em 9th International Conference on Learning Representations}, 2021.

\bibitem{su2022dual}
Xuan Su, Jiaming Song, Chenlin Meng, and Stefano Ermon.
\newblock Dual diffusion implicit bridges for image-to-image translation.
\newblock In {\em The Eleventh International Conference on Learning Representations}, 2023.

\bibitem{tang2023emergent}
Luming Tang, Menglin Jia, Qianqian Wang, Cheng~Perng Phoo, and Bharath Hariharan.
\newblock Emergent correspondence from image diffusion.
\newblock In {\em Annual Conference on Neural Information Processing Systems}, 2023.

\bibitem{taunk2019brief}
Kashvi Taunk, Sanjukta De, Srishti Verma, and Aleena Swetapadma.
\newblock A brief review of nearest neighbor algorithm for learning and classification.
\newblock In {\em 2019 international conference on intelligent computing and control systems (ICCS)}, pages 1255--1260. IEEE, 2019.

\bibitem{tian2023diffuse}
Junjiao Tian, Lavisha Aggarwal, Andrea Colaco, Zsolt Kira, and Mar Gonz{\'{a}}lez{-}Franco.
\newblock Diffuse, attend, and segment: Unsupervised zero-shot segmentation using stable diffusion.
\newblock {\em CoRR}, abs/2308.12469, 2023.

\bibitem{vaswani2017attention}
Ashish Vaswani, Noam Shazeer, Niki Parmar, Jakob Uszkoreit, Llion Jones, Aidan~N. Gomez, Lukasz Kaiser, and Illia Polosukhin.
\newblock Attention is all you need.
\newblock In {\em Annual Conference on Neural Information Processing Systems}, pages 5998--6008, 2017.

\bibitem{Vilas23Analyzing}
Martina~G. Vilas, Timothy Schauml{\"{o}}ffel, and Gemma Roig.
\newblock Analyzing vision transformers for image classification in class embedding space.
\newblock In {\em Annual Conference on Neural Information Processing Systems 2023}, 2023.

\bibitem{wan2023harnessing}
Qiang Wan, Zilong Huang, Bingyi Kang, Jiashi Feng, and Li~Zhang.
\newblock Harnessing diffusion models for visual perception with meta prompts.
\newblock {\em CoRR}, abs/2312.14733, 2023.

\bibitem{DBLP:conf/nips/WangX00CH22}
Zitai Wang, Qianqian Xu, Zhiyong Yang, Yuan He, Xiaochun Cao, and Qingming Huang.
\newblock Openauc: Towards auc-oriented open-set recognition.
\newblock In {\em Annual Conference on Neural Information Processing Systems}, 2022.

\bibitem{DBLP:journals/pami/WangXYHCH23}
Zitai Wang, Qianqian Xu, Zhiyong Yang, Yuan He, Xiaochun Cao, and Qingming Huang.
\newblock Optimizing partial area under the top-k curve: Theory and practice.
\newblock {\em {IEEE} Trans. Pattern Anal. Mach. Intell.}, 45(4):5053--5069, 2023.

\bibitem{DBLP:conf/nips/WangX00CH23}
Zitai Wang, Qianqian Xu, Zhiyong Yang, Yuan He, Xiaochun Cao, and Qingming Huang.
\newblock A unified generalization analysis of re-weighting and logit-adjustment for imbalanced learning.
\newblock In {\em Annual Conference on Neural Information Processing Systems}, 2023.

\bibitem{DBLP:conf/nips/WuZCGZHZSS23}
Weijia Wu, Yuzhong Zhao, Hao Chen, Yuchao Gu, Rui Zhao, Yefei He, Hong Zhou, Mike~Zheng Shou, and Chunhua Shen.
\newblock Datasetdm: Synthesizing data with perception annotations using diffusion models.
\newblock In {\em Annual Conference on Neural Information Processing Systems}, 2023.

\bibitem{xiao2023text}
Changming Xiao, Qi~Yang, Feng Zhou, and Changshui Zhang.
\newblock From text to mask: Localizing entities using the attention of text-to-image diffusion models.
\newblock {\em CoRR}, abs/2309.04109, 2023.

\bibitem{xu2023open}
Jiarui Xu, Sifei Liu, Arash Vahdat, Wonmin Byeon, Xiaolong Wang, and Shalini~De Mello.
\newblock Open-vocabulary panoptic segmentation with text-to-image diffusion models.
\newblock In {\em {IEEE/CVF} Conference on Computer Vision and Pattern Recognition}, pages 2955--2966, 2023.

\bibitem{yang2023diffusion}
Xingyi Yang and Xinchao Wang.
\newblock Diffusion model as representation learner.
\newblock In {\em {IEEE/CVF} International Conference on Computer Vision}, pages 18892--18903, 2023.

\bibitem{yu15lsun}
Fisher Yu, Yinda Zhang, Shuran Song, Ari Seff, and Jianxiong Xiao.
\newblock {LSUN:} construction of a large-scale image dataset using deep learning with humans in the loop.
\newblock {\em CoRR}, abs/1506.03365, 2015.

\bibitem{DBLP:journals/corr/abs-2308-06739}
David~Junhao Zhang, Mutian Xu, Chuhui Xue, Wenqing Zhang, Xiaoguang Han, Song Bai, and Mike~Zheng Shou.
\newblock Free-atm: Exploring unsupervised learning on diffusion-generated images with free attention masks.
\newblock {\em CoRR}, abs/2308.06739, 2023.

\bibitem{zhang2023tale}
Junyi Zhang, Charles Herrmann, Junhwa Hur, Luisa~Polania Cabrera, Varun Jampani, Deqing Sun, and Ming{-}Hsuan Yang.
\newblock A tale of two features: Stable diffusion complements {DINO} for zero-shot semantic correspondence.
\newblock In {\em Annual Conference on Neural Information Processing Systems}, 2023.

\bibitem{Zhang18ResUnet}
Zhengxin Zhang, Qingjie Liu, and Yunhong Wang.
\newblock Road extraction by deep residual u-net.
\newblock {\em {IEEE} Geosci. Remote. Sens. Lett.}, 15(5):749--753, 2018.

\bibitem{zhao2023unleashing}
Wenliang Zhao, Yongming Rao, Zuyan Liu, Benlin Liu, Jie Zhou, and Jiwen Lu.
\newblock Unleashing text-to-image diffusion models for visual perception.
\newblock In {\em {IEEE/CVF} International Conference on Computer Vision}, pages 5706--5716, 2023.

\bibitem{Zhou17ade20k}
Bolei Zhou, Hang Zhao, Xavier Puig, Sanja Fidler, Adela Barriuso, and Antonio Torralba.
\newblock Scene parsing through {ADE20K} dataset.
\newblock In {\em {IEEE} Conference on Computer Vision and Pattern Recognition}, pages 5122--5130, 2017.

\bibitem{zhou2023hyperspectral}
Jingyi Zhou, Jiamu Sheng, Jiayuan Fan, Peng Ye, Tong He, Bin Wang, and Tao Chen.
\newblock When hyperspectral image classification meets diffusion models: An unsupervised feature learning framework.
\newblock {\em CoRR}, abs/2306.08964, 2023.

\end{thebibliography}


\newpage
\appendix

\section*{\Large{Contents}}
\startcontents[appendices]
\printcontents[appendices]{l}{1}{\setcounter{tocdepth}{3}}
\newpage

\section{Additional Visualization for Distinct Properties of Diffusion U-Nets}
\label{sec:app-visualization}
\subsection{Visualization of Other Sample Images}

\begin{figure}[h]
  \centering
  \includegraphics[width=.9\linewidth]{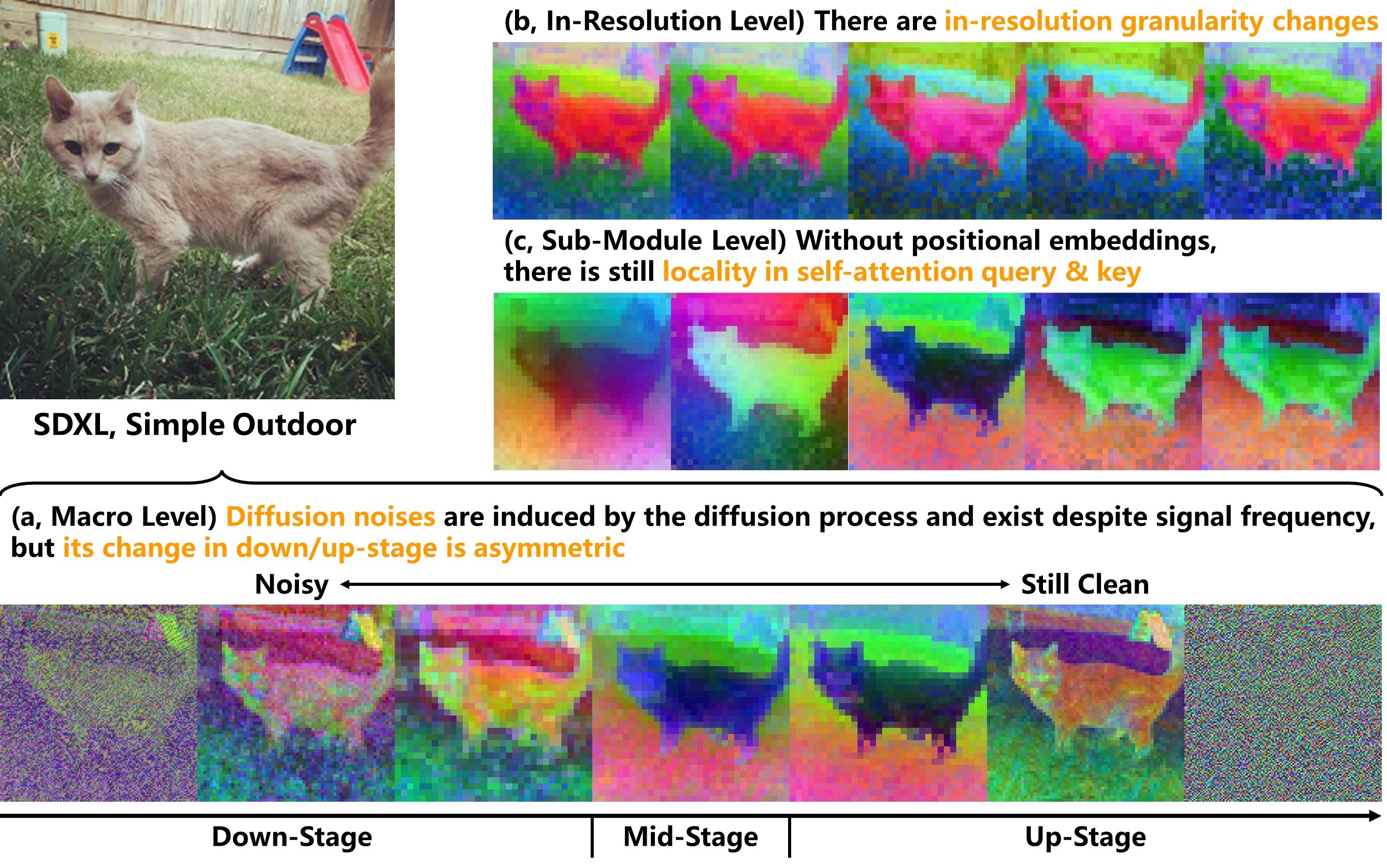}
  \caption{Visualization of SDXL activations on a simple outdoor scene.}
  \label{fig:app-sdxl-cat}
\end{figure}

\begin{figure}[h]
  \centering
  \includegraphics[width=.9\linewidth]{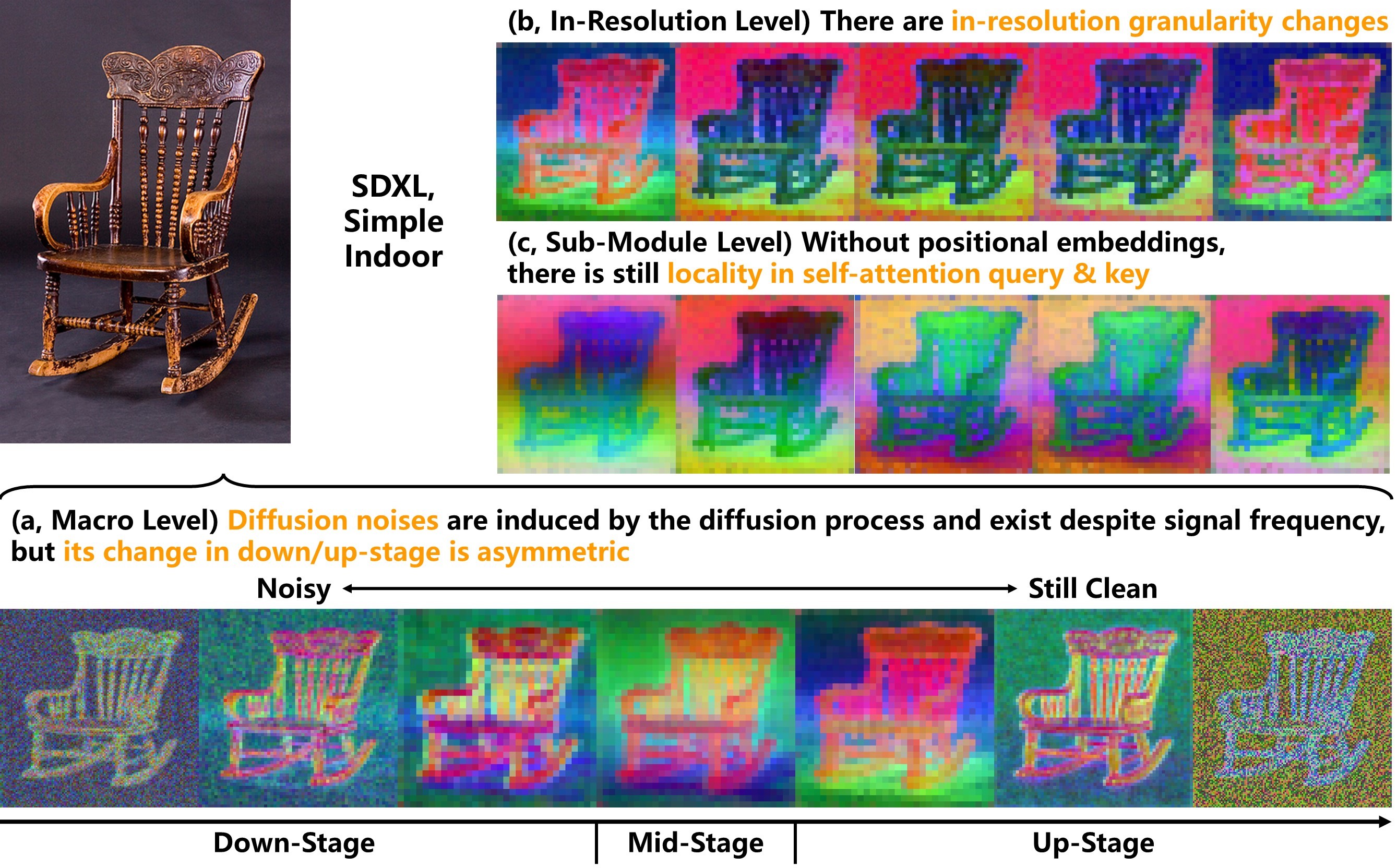}
  \caption{Visualization of SDXL activations on a simple indoor scene.}
  \label{fig:app-sdxl-chair}
\end{figure}

\begin{figure}[t]
  \centering
  \includegraphics[width=.9\linewidth]{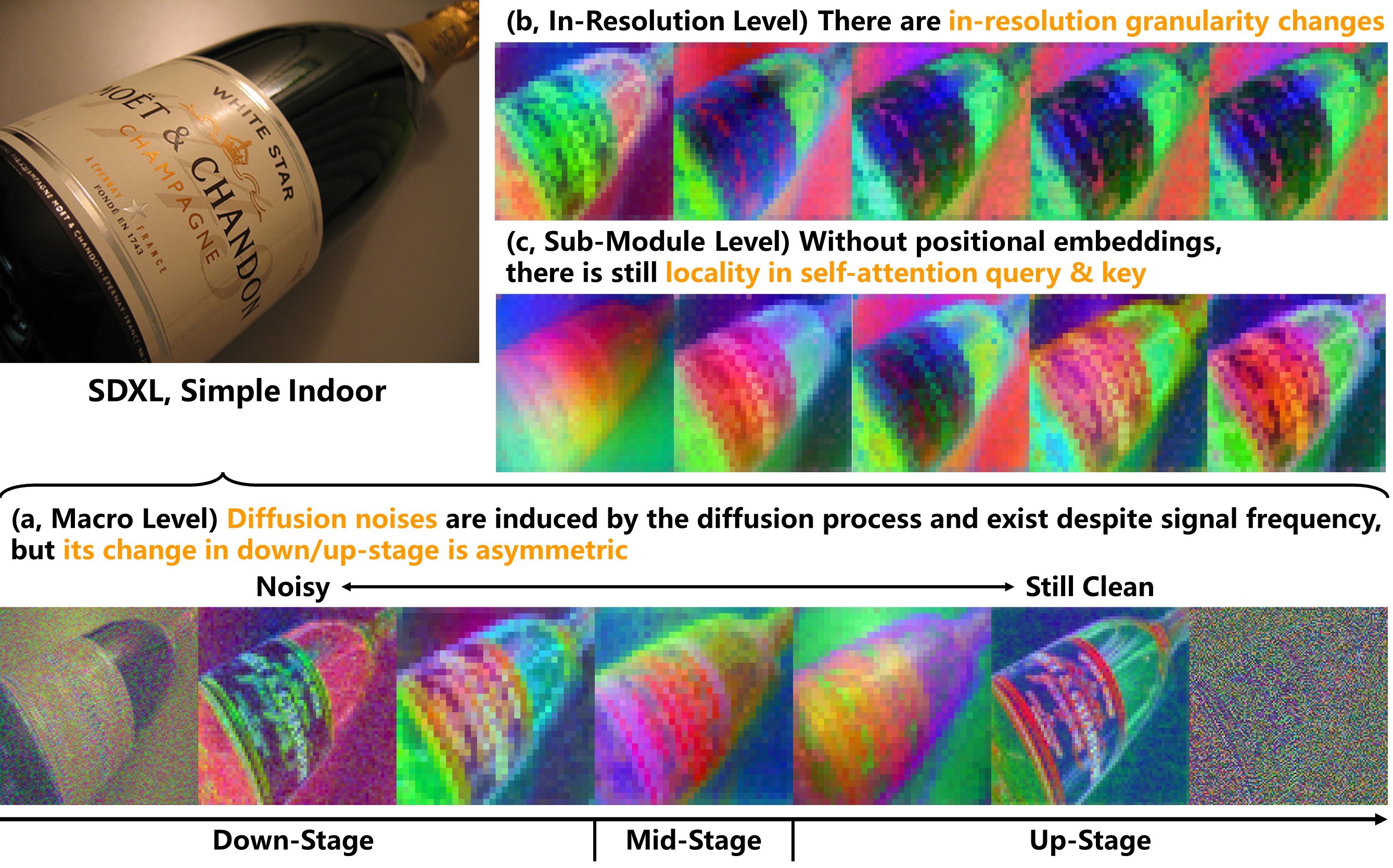}
  \caption{Visualization of SDXL activations on a simple indoor scene.}
  \label{fig:app-sdxl-bottle}
\end{figure}

\begin{figure}[t]
  \centering
  \includegraphics[width=.9\linewidth]{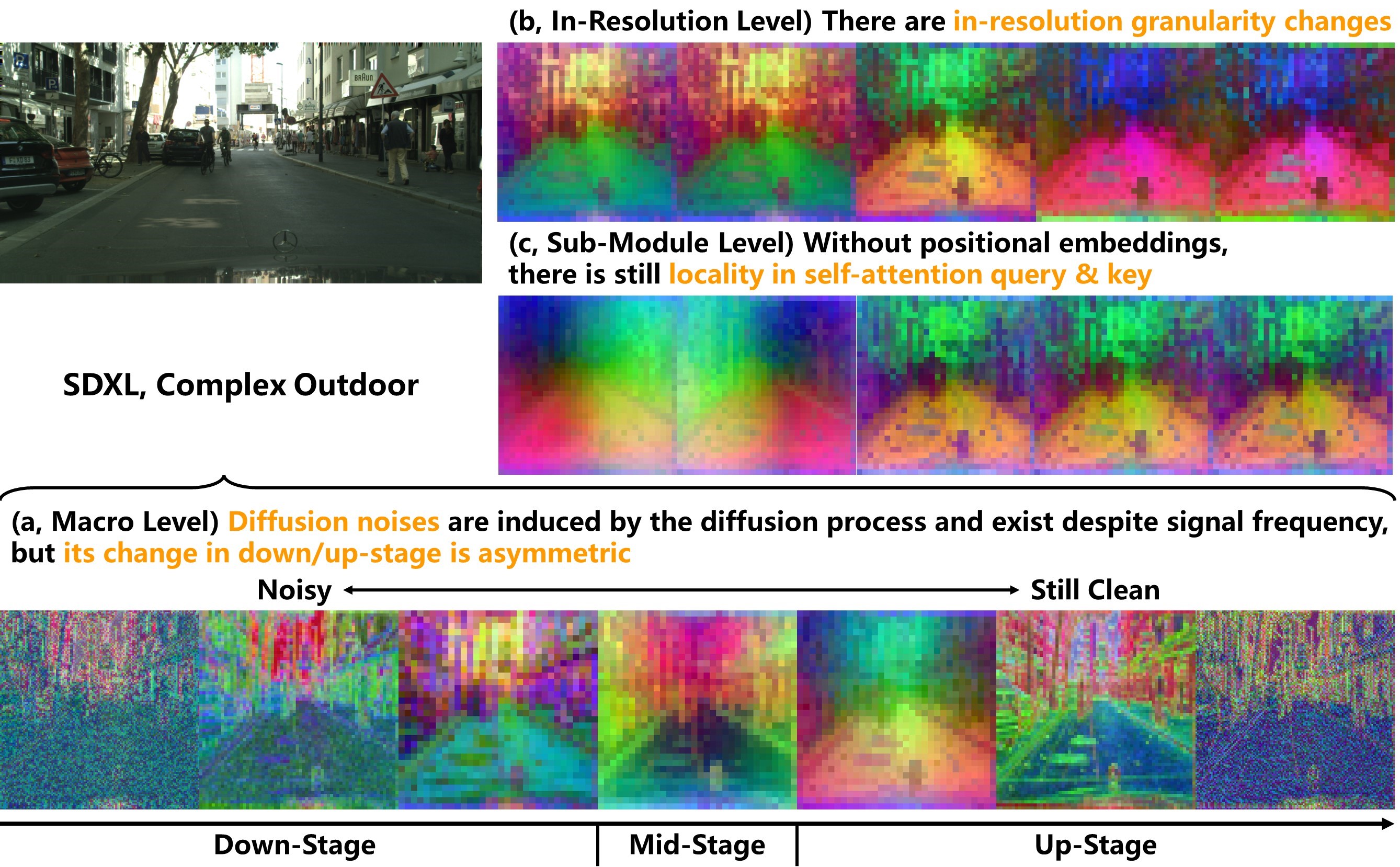}
  \caption{Visualization of SDXL activations on a complex outdoor scene.}
  \label{fig:app-sdxl-city}
\end{figure}

\begin{figure}[t]
  \centering
  \includegraphics[width=.9\linewidth]{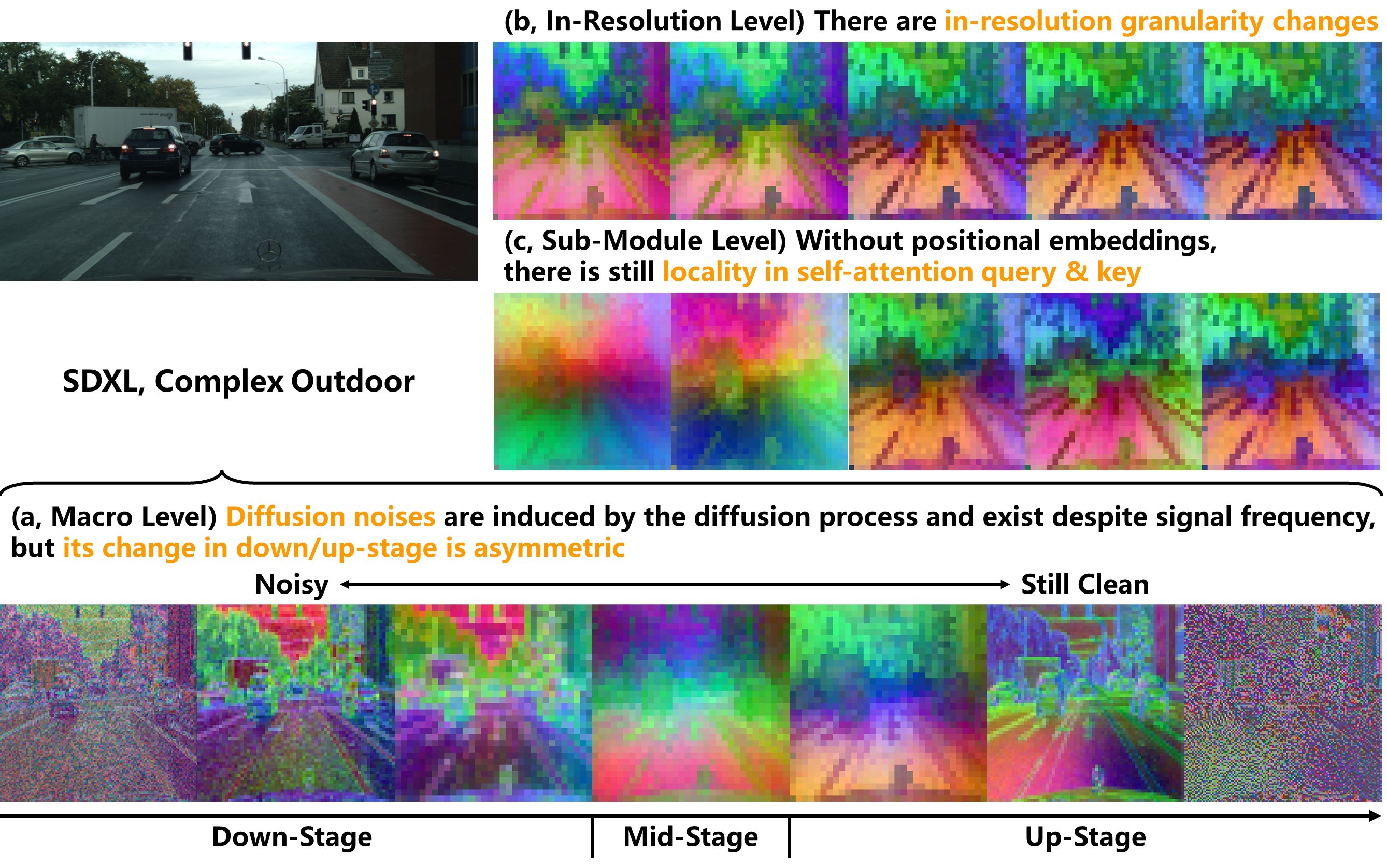}
  \caption{Visualization of SDXL activations on a complex outdoor scene.}
  \label{fig:app-sdxl-city2}
\end{figure}

\begin{figure}[t]
  \centering
  \includegraphics[width=.9\linewidth]{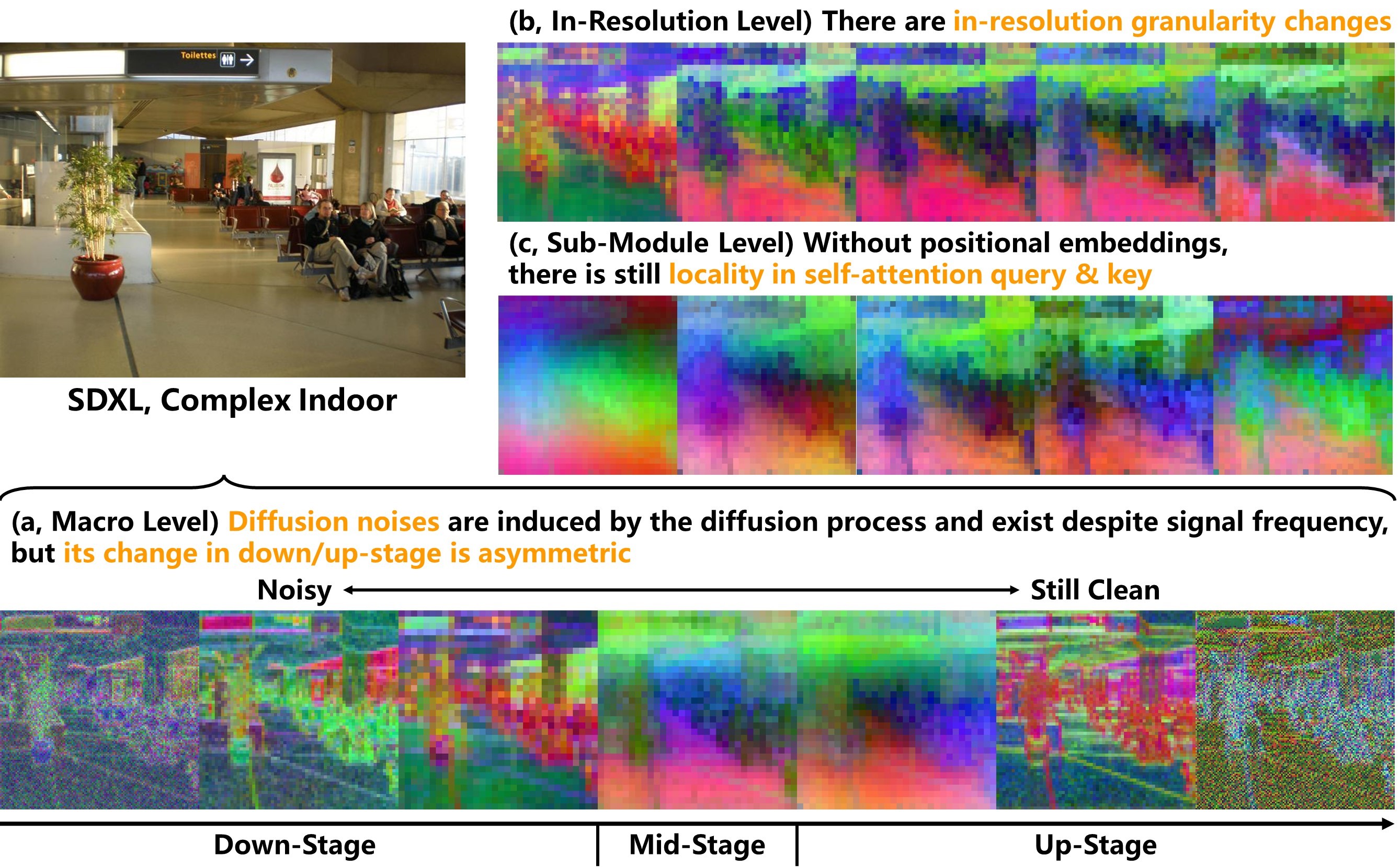}
  \caption{Visualization of SDXL activations on a complex indoor scene.}
  \label{fig:app-sdxl-ade}
\end{figure}

\begin{figure}[t]
  \centering
  \includegraphics[width=.9\linewidth]{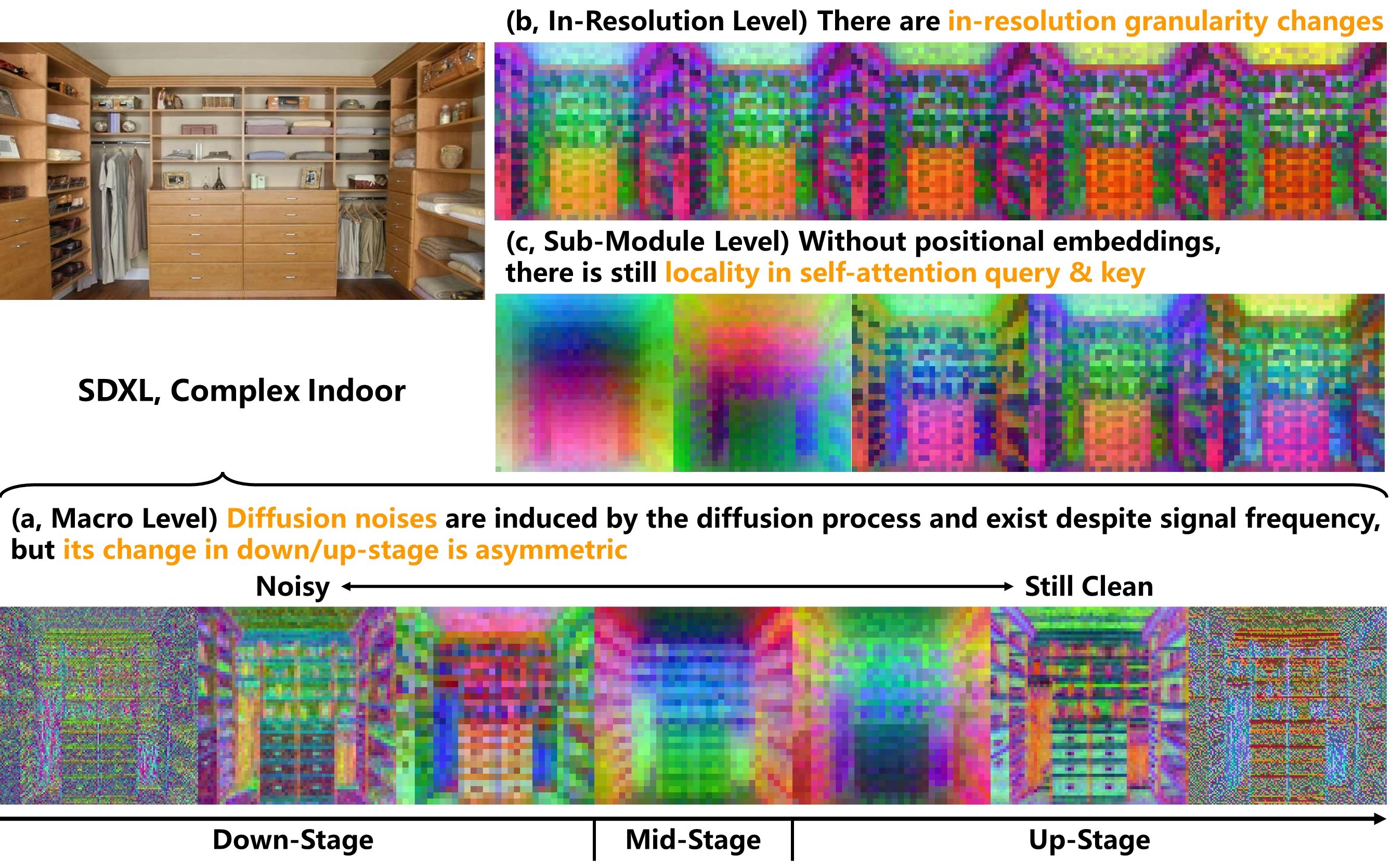}
  \caption{Visualization of SDXL activations on a complex indoor scene.}
  \label{fig:app-sdxl-ade2}
\end{figure}

We have provided activation visualization from SDXL in a simple outdoor scene presenting a horse in the wild.
Next, we will show visualization from the same model in different scenes to demonstrate the universality of the observed properties.
Another simple outdoor scene presenting a cat is in \Cref{fig:app-sdxl-cat}.
Two simple indoor scenes are visualized in \Cref{fig:app-sdxl-chair} and \Cref{fig:app-sdxl-bottle}.
Two complex outdoor scenes of urban streets are visualized in \Cref{fig:app-sdxl-city} and \Cref{fig:app-sdxl-city2}.
Two complex indoor scenes are shown in \Cref{fig:app-sdxl-ade} and \Cref{fig:app-sdxl-ade2}.

\subsection{Visualization from Other Diffusion Models}
\begin{figure}[t]
  \centering
  \includegraphics[width=.9\linewidth]{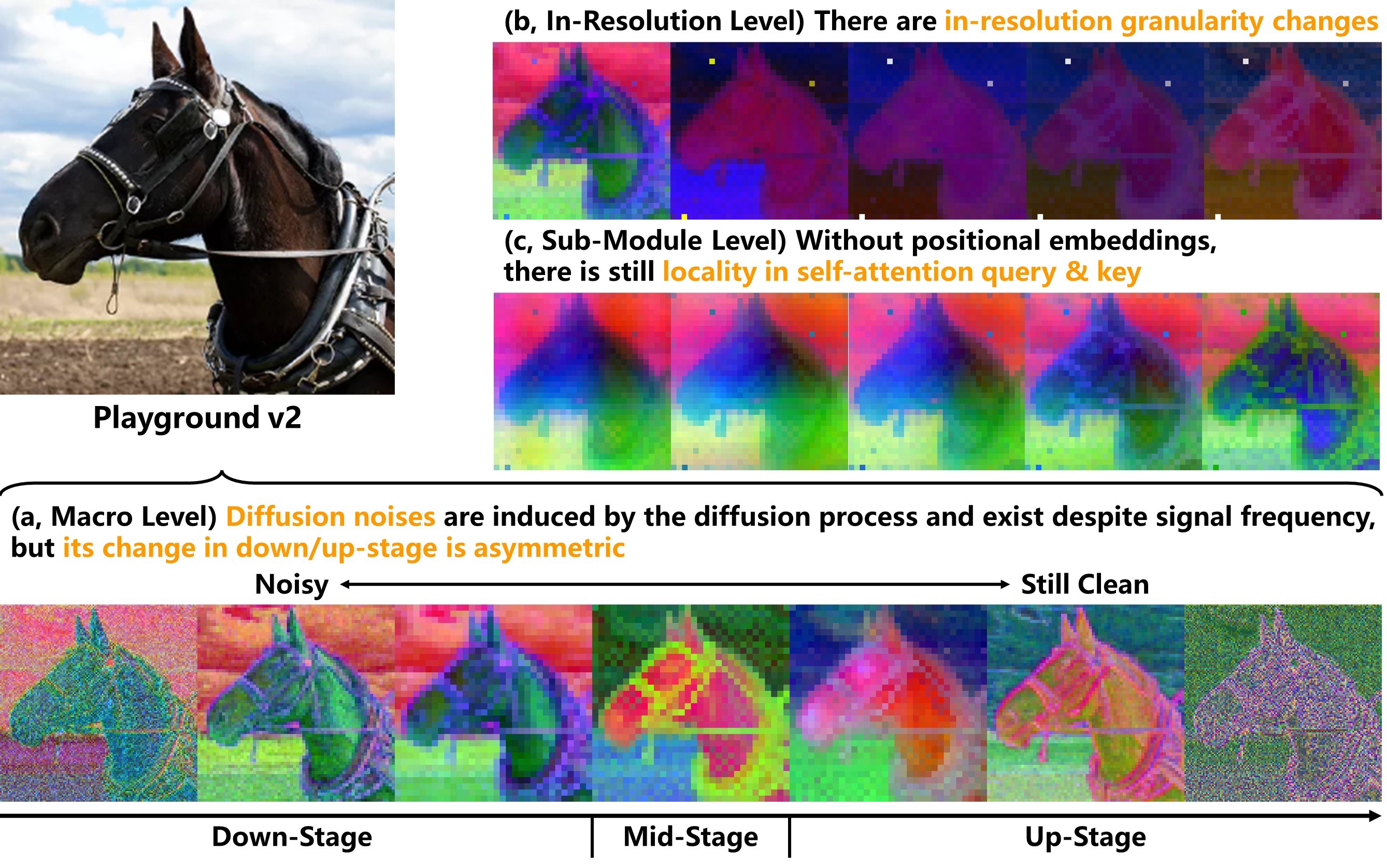}
  \caption{Visualization of Playground v2 activations on a simple outdoor scene.}
  \label{fig:app-pgv2-horse}
\end{figure}

\textbf{Playground v2 Activations.}
Playground v2\footnote{\url{https://huggingface.co/playgroundai/playground-v2-1024px-aesthetic}} shares the same architecture as SDXL but is trained independently, and it is claimed to be more powerful in generation.
In \Cref{fig:app-pgv2-horse}, we present its activation visualization using the same horse image as the primary SDXL visualization.
Compared to SDXL, Playground v2 activations are less noisy, particularly in the down-stage.
This supports the claim that Playground v2 is a stronger model.

\begin{figure}[t]
  \centering
  \includegraphics[width=.9\linewidth]{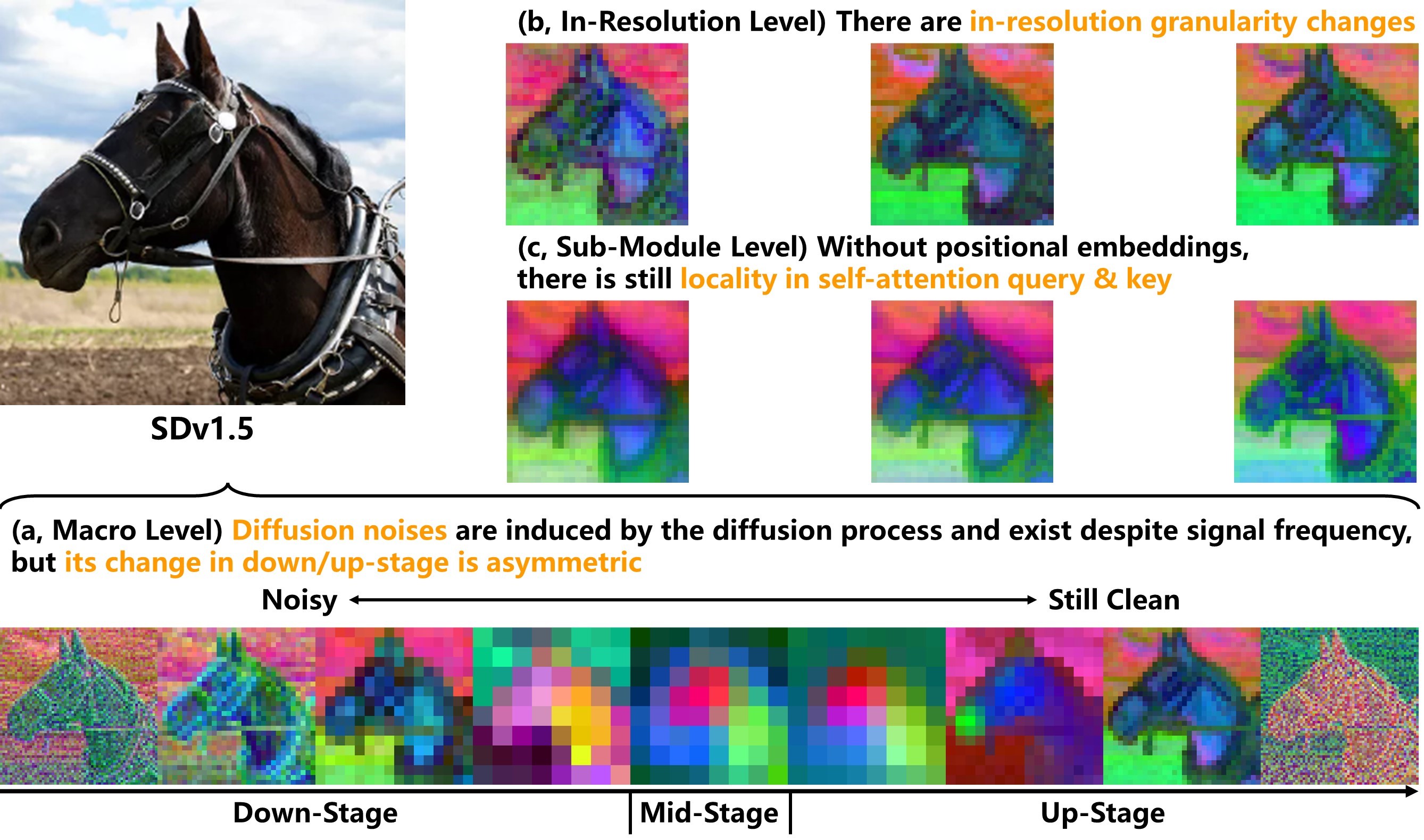}
  \caption{Visualization of SDv1.5 activations on a simple outdoor scene.}
  \label{fig:app-sd15-horse}
\end{figure}

\textbf{SDv1.5 Activations.}
SDv1.5, as an older model, has a slightly different architecture from SDXL.
Specifically, SDv1.5 has four resolutions instead of three, but its ViTs contain only one layer.
Moreover, in SDXL, only the highest resolution lacks ViT due to efficiency concerns, while in SDv1.5, only the lowest resolution lacks ViT due to its low resolution.
Despite the architectural changes, the three unique properties still apply to SDv1.5, as shown in \Cref{fig:app-sd15-horse}.

\begin{figure}[t]
  \centering
  \includegraphics[width=.9\linewidth]{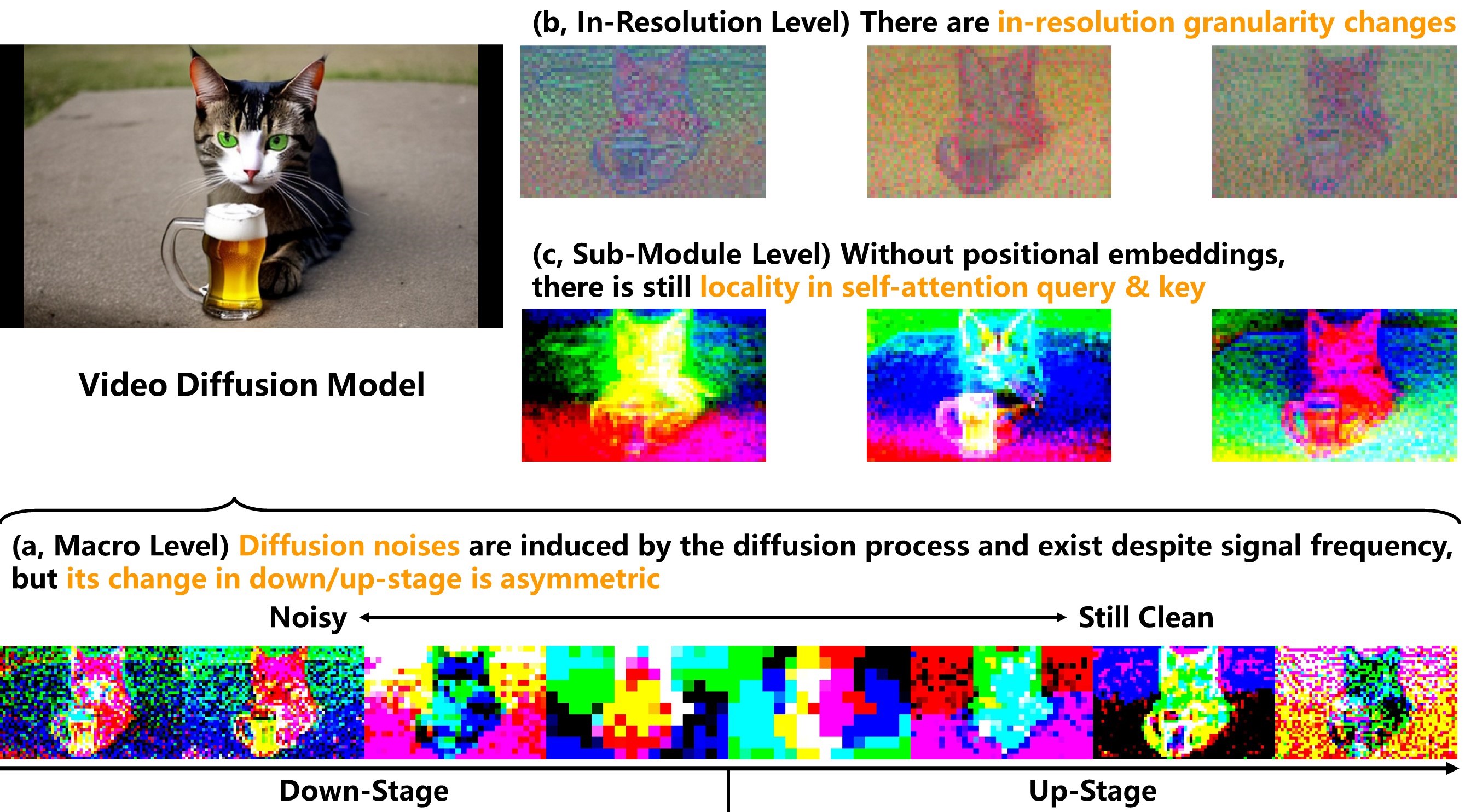}
  \caption{Visualization of the activations of a video diffusion model. The input is a short video, and we visualize one frame of it.}
  \label{fig:app-video}
\end{figure}

\textbf{Video Diffusion Activations.}
To further demonstrate the universality, we even select a diffusion model for video generation~\cite{chen2024videocrafter2} for visualization.
This model is based on the SDv1.5 architecture, with an additional temporal attention layer inserted after cross-attention to enable sequential generation.
Although this model is designed for a different task, we can still observe the three unique properties in \Cref{fig:app-video}.

\begin{figure}[t]
  \centering
  \includegraphics[width=.5\linewidth]{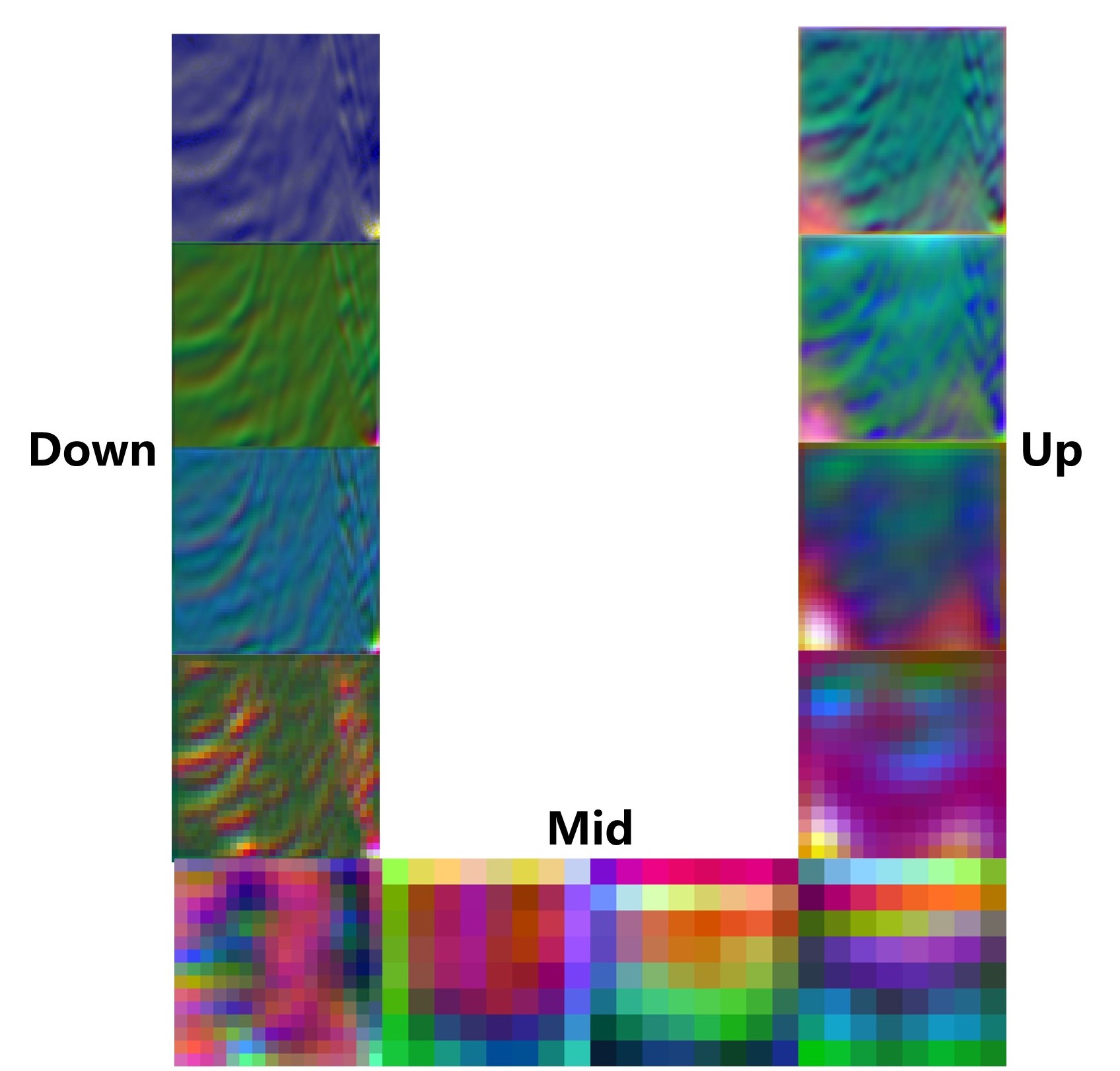}
  \caption{Activation visualization from a traditional U-Net for semantic segmentation~\cite{Zhang18ResUnet}.}
  \label{fig:app-unet}
\end{figure}

\textbf{Conventional U-Net Activations.}
As comparison to diffusion U-Net, we also visualize some activations from a conventional U-Net in \Cref{fig:app-unet}, where the three properties of diffusion U-Net can hardly be observed.

\clearpage


\section{Details of Our Method}
\label{sec:app-selection}
In this section, we index all diffusion U-Net components in the order in which they are activated during a network forward run.
Besides descriptions in natural language, we also index the activations using the same notations as used in our code implementation.
These notations denote stages using \textit{down/mid/up}, resolutions as \textit{level}, and module index as \textit{repeat}.
Index starts to count from 0 instead of 1, following the convention in coding.

\subsection{Feature Selection Solution for SDv1.5}
We select four activations from SDv1.5, maintaining the same total feature channels as the conventional way to extract all output activations from each resolution.
\begin{enumerate}[label=(\roman*), leftmargin=25pt]
    \item The cross-attention query activation from the 2nd ViT, the 2nd resolution. This provides coarse information for simple scenes (up-level1-repeat1-vit-block0-cross-q).
    \item The inter-module activation after the 3rd ResModule, the 2nd resolution. This provides coarse information for complex scenes (up-level1-repeat2-res-out).
    \item The cross-attention query activation from the 2nd ViT, the 3rd resolution. This provides finer information with a higher resolution (up-level2-repeat1-vit-block0-cross-q).
    \item The self-attention key activation from the 1st ViT, the 4th resolution. This extracts features from the highest resolution, harnessing the noise suppression effect of self-attention locality (up-level3-repeat0-vit-block0-self-k).
\end{enumerate}
We omit the index of basic blocks in ViTs, as SDv1.5 only contains one-layer ViTs.
Additionally, we totally ignore the lowest resolution, as its activations have very low resolution ($8\times8$ if the input image is $512\times512$), suggesting inferiority, as supported by the quantitative comparison.

\subsection{Feature Selection Solution for SDXL}
Four activations are selected from SDXL, trying to get similar total feature channels to the SDv1.5 feature selection solution.
\begin{enumerate}[label=(\roman*), leftmargin=25pt]
    \item The output activation after the 8th basic block, the 1st ViT, the 1st resolution. This provides relatively coarse information for simple scenes (up-level0-repeat0-vit-block7-out).
    \item The output activation after the 6th basic block, the 1st ViT, the 1st resolution. This provides relatively coarse information for complex scenes (up-level0-repeat0-vit-block5-out).
    \item The cross-attention query activation from the 1st basic block, the 1st ViT, the 2nd resolution. This provides relatively fine information for simple scenes (up-level1-repeat0-vit-block0-cross-q).
    \item The output activation after the 1st basic block, the 1st ViT, the 2nd resolution. This provides relatively fine information for complex scenes (up-level1-repeat0-vit-block0-out).
\end{enumerate}
We ignore the highest resolution since it is affected by diffusion noises and lacks ViTs from which we can extract self-attention activations.
However, if the downstream model is strong enough to learn to suppress noises, it may be possible to additionally extract inter-module activations from this resolution to harness more information.

\subsection{Feature Selection Solution with Additional Techniques}
For the setting \textit{Ours-XL-t}, we mainly utilize two simple techniques that are also adopted in some SOTA methods:
(i) We additionally extract attention maps, \textit{i.e.}, the similarity scores of cross-attention query and key, as dense features~\cite{zhao2023unleashing, xu2023open}.
Such attention maps are closely related to the semantics of prompts, thus providing important supplementary information.
Despite its usefulness, this technique only adds a few additional channels to the features.
(ii) We amalgamate features from different models~\cite{baranchuk2021label, luo2023dhf, zhou2023hyperspectral} through simple concatenation.
This technique is a common practice and can be seen as an extension of amalgamating different activations together as a whole feature.
We next explain what activations are selected to implement the two techniquess.

We first extract features according to the feature selection solution for SDXL.
Afterward, we extract additional features from SDv1.5:
\begin{enumerate}[label=(\roman*), leftmargin=25pt]
    \item The cross-attention query activation from the 2nd ViT, the 2nd resolution (up-level1-repeat1-vit-block0-cross-q).
    \item The cross-attention query activation from the 2nd ViT, the 3rd resolution (up-level2-repeat1-vit-block0-cross-q).
    \item The upsampler output activation from the 3rd resolution, to harness high-resolution information (up-level2-upsampler-out).
    \item The self-attention key activation from the 1st ViT, the 4th resolution, to harness high-resolution information (up-level3-repeat0-vit-block0-self-k).
    \item The attention maps averaged over all cross-attention layers in the up-stage.
\end{enumerate}
We also extract one feature from Playground v2:
the output activation after the 4th basic block, the 1st ViT, the 1st resolution (up-level0-repeat0-vit-block3-out).

\subsection{Alternative Feature Selection Solution with Additional Techniques}
Large-scale datasets for semantic segmentation mostly consist of images of complex scenes.
In such cases, we find attention maps can be too noisy to be useful.
Therefore, we discard the attention map technique and the entire SDv1.5 model, as it is weaker compared to the newer SDXL and Playground v2 models.
To compensate for the loss of activations, we select additional activations from SDXL and Playground v2.

From SDXL, we select all the activations as described in the feature selection solution for SDXL and select one additional activation:
the upsampler output activation from the 2nd resolution, to harness high-resolution information (up-level1-upsampler-out).
From Playground v2, we select the following activations:
\begin{enumerate}[label=(\roman*), leftmargin=25pt]
    \item The output activation after the 6th basic block, the 1st ViT, the 1st resolution (up-level0-repeat0-vit-block5-out).
    \item The cross-attention query activation from the 1st basic block, the 1st ViT, the 2nd resolution (up-level1-repeat0-vit-block0-cross-q).
    \item The upsampler output activation from the 2nd resolution (up-level1-upsampler-out).
\end{enumerate}


\section{Experimental Details}
\label{sec:app-detail}
\subsection{Semantic Correspondence}

\textbf{Task and Dataset.}
Semantic correspondence~\cite{hao2020brief} involves finding a pixel in an image that semantically matches another keypoint pixel in a reference image, such as the hind legs of two different cats.
We conduct experiments on the SPair-71k dataset~\cite{min2019spair}.

\textbf{Evaluation Metric.}
$\text{PCK@0.1}_{\text{img}}(\uparrow)$ and $\text{PCK@0.1}_{\text{bbox}}(\uparrow)$ are used, following the widely-adopted protocol reported in \cite{min2019spair}. These two metrics mean the percentage of correctly predicted keypoints, where a predicted keypoint is considered to be correct if it lies within the neighborhood of the corresponding annotation with a radius of $0.1 \times max(h, w)$. For $\text{PCK@0.1}_{\text{img}}$/$\text{PCK@0.1}_{\text{bbox}}$, $h, w$ denote the dimension of the entire image/object bounding box, respectively.

\textbf{SOTA Competitors.}
We provide the results from four SOTA methods: DINO~\cite{dino} and DHPF~\cite{min2020learning} as non-diffusion-feature methods, as well as DIFT~\cite{tang2023emergent} and DHF~\cite{luo2023dhf} as diffusion feature methods.

\textbf{Implementation Details.}
The semantic correspondence task can be done via the nearest neighbor algorithm~\cite{taunk2019brief}, which is unsupervised and training-free~\cite{min2019spair}.
We add one additional trainable convolutional layer before applying the nearest neighbor algorithm to refine the input features, which is also adopted by some SOTAs including DHF.
The model is trained for two epochs, each containing 5,000 sample pairs, following conventional settings.
Our implementation is derived from DHF, and we keep all hyper-parameters at their default settings.


\subsection{Semantic Segmentation}

\textbf{Task and Dataset.}
Semantic segmentation~\cite{hao2020brief} is essentially pixel-level classification.
For this task, we choose the ADE20K dataset~\cite{Zhou17ade20k} with over 20k annotated images of 150 semantic categories, and the CityScapes dataset~\cite{Cordts16city}, which contains 5,000 fine annotated images of urban street scenes.

\textbf{Evaluation Metric.}
We use mIoU metric, which is the mean over the IoU performance across all semantic classes \cite{hao2020brief}. For each image, IoU (Intersection over Union, $\uparrow$) is defined by \#(overlapped pixels between the prediction and the ground truth) / \#(union pixels of them).

\textbf{SOTA Competitors.}
We choose three diffusion feature SOTA methods as competitors.
ODISE~\cite{xu2023open} is an early method with a simple implementation.
VPD~\cite{zhao2023unleashing} is another early study of this field, which introduces additional text adapter modules for improvement.
Meta Prompts~\cite{wan2023harnessing} is a newer method and shows significant improvements.
We also report the performance of MaskCLIP~\cite{Dong23maskclip}, which is included as a competitor in the ODISE study.

\textbf{Implementation Details.}
Both VPD and Meta Prompts perform full-scale fine-tuning on the diffusion U-Net using feedback from the discriminative task.
This heavy fine-tuning does not entirely comply with the motivation of diffusion feature, which seeks a balance between wider applicability and less training, and is hard to extend to the larger SDXL model.
Therefore, we keep the entire diffusion model frozen instead.
As the setting has been changed, the performance of VPD and Meta Prompts in \Cref{tab:exp-segm} is based on our experiments, not the reported results from the original papers.
Our implementation directly uses the hyper-parameters reported in Meta Prompts.


\subsection{Label-Scarce Segmentation}

\textbf{Task and Dataset.}
Using features from a pre-trained diffusion model ensures good performance even when labeled training data is scarce~\cite{baranchuk2021label}.
For this setting, we use a dataset collected in \cite{baranchuk2021label} and experiment on its Horse-21 subset, the data of which is sourced from LSUN~\cite{yu15lsun}.
This subset contains only 30 labeled training images to be consistent with the intuition.
The semantic segmentation in the label-scarce scenario also uses the mIoU metric.

\textbf{SOTA Competitors.}
We select the SOTA diffusion feature approach, DDPM~\cite{baranchuk2021label}, as the major competitor.
We also include other representative segmentation methods: DatasetDDPM, MAE~\cite{he2022masked}, SwAV~\cite{caron2020unsupervised}, which are all reported in \cite{baranchuk2021label}.

\textbf{Implementation Details.}
Following DDPM~\cite{baranchuk2021label}, the downstream model is an ensemble of ten simple MLP networks, each conducting pixel-wise classification.
The simplicity of the model is intended to demonstrate the innate capability and generalizability of diffusion models.
Our implementation is derived from DDPM with only batch size changed among all hyper-parameters.
We use a larger batch size for faster experiments as a smaller one does not improve performance.
\textbf{Additionally, this is also the setting for the quantitative comparison, as the compact size of the dataset can enhance efficiency}.


\section{Additional Experimental Results}
\label{sec:app-exp}

\subsection{Generalizability across Different Scenes}
As stated in \Cref{sec:method}, it is preferable to conduct the quantitative comparison across multiple datasets and choose activations that are optimal for each.
This approach can enhance the generalizability of the selected features.
To evaluate the generalizability of our features, we conducted an additional experiment, with results presented in \Cref{tab:app-exp2}.

In this experiment, we design an alternative feature selection solution for SDXL, based solely on quantitative results from a single dataset consisting of simple scenes.
In this solution, we extract both optimal and slightly sub-optimal activations, maintaining the same total number of feature channels as the standard solution.
The alternative solution achieves higher performance on the simple scene it is based on but performs significantly worse on the other scene.
Therefore, we conclude that our standard feature selection solution achieves generalizability across different scenes, albeit with a slight performance drop compared to features specifically selected for each scene.

\begin{table}[t]
\centering
\caption{Examination of generalizability across different scenes. \textit{Generic Solution} refers to our standard feature selection solution for SDXL, while \textit{Specific Solution} refers to the outcome of considering only a simple scene for the quantitative comparison. The experiment is conducted on the label-scarce segmentation task, where the Horse-21 subset is used for simple scenes and the Bedroom-28 subset is used for complex scenes.
We mark the better results as \textcolor{orange}{\textbf{bold}} font.}
\label{tab:app-exp2}
\begin{tabular}{@{}lcc@{}}
\toprule
Method             & Simple Scene & Complex Scene \\ \midrule
Generic Solution   & 63.34        & \textcolor{orange}{\textbf{47.55}}         \\
Specific Solution & \textcolor{orange}{\textbf{63.70}}        & 45.41         \\ \bottomrule
\end{tabular}
\end{table}

\subsection{Quantitative Comparison Results}
\begin{table}[t]
\centering
\caption{Quantitative comparison results of SDXL and Playground v2. This table shows the results from the lowest resolution.
Activation ID indicates the location of each activation in the diffusion U-Net.
The best results are in \textcolor{orange}{\textbf{bold}} font and the runner-up is \textblue{\underline{underlined}}.}
\label{tab:app-comparison-1}
\begin{tabular}{@{}lcccc@{}}
\toprule
\multirow{2}{*}{Activation   ID}   & \multicolumn{2}{c}{SDXL}        & \multicolumn{2}{c}{Playground v2} \\
                                   & Simple      & Complex     & Simple & Complex     \\ \midrule
up-level0-repeat0-res-out            & 53.49       & 38.52       & 54.57  & 40.73       \\
up-level0-repeat0-vit-block0-cross-q & 56.46       & 42.31       & 55.32  & 41.93       \\
up-level0-repeat0-vit-block0-out     & 57.36       & 42.31       & 56.60  & 44.52       \\
up-level0-repeat0-vit-block1-cross-q & 57.82       & 42.72       & 56.44  & 43.93       \\
up-level0-repeat0-vit-block1-out     & \textblue{\underline{58.73}} & 42.62       & 58.12  & 45.82       \\
up-level0-repeat0-vit-block3-cross-q & 58.27       & 43.09       & 57.79  & 45.90       \\
up-level0-repeat0-vit-block3-out     & \textcolor{orange}{\textbf{58.95}} & \textblue{\underline{44.83}}    & \textcolor{orange}{\textbf{59.04}}  & 47.81           \\
up-level0-repeat0-vit-block5-cross-q & 57.30       & 44.36       & 56.97  & 47.38       \\
up-level0-repeat0-vit-block5-out     & 58.70          & \textcolor{orange}{\textbf{45.26}} & 58.67           & \textcolor{orange}{\textbf{49.77}}  \\
up-level0-repeat0-vit-block7-cross-q & 57.24       & 43.00       & 57.69  & 48.51       \\
up-level0-repeat0-vit-block7-out     & \textcolor{orange}{\textbf{58.95}} & \textblue{\underline{44.83}}    & \textblue{\underline{58.76}}     & 48.39           \\
up-level0-repeat0-vit-block9-cross-q & 57.06       & 41.73       & 56.03  & 44.68       \\
up-level0-repeat0-vit-block9-out     & 58.46       & 43.98       & 56.59  & \textblue{\underline{48.54}} \\
up-level0-repeat0-vit-out            & 58.36       & 41.54       & 57.99  & 43.95       \\
up-level0-repeat1-res-out            & 57.28       & 39.27       & 56.59  & 41.74       \\
up-level0-repeat1-vit-block0-cross-q & 55.72       & 40.96       & 55.56  & 43.97       \\
up-level0-repeat1-vit-block0-out     & 56.70       & 40.97       & 57.51  & 43.35       \\
up-level0-repeat1-vit-block1-cross-q & 57.92       & 41.65       & 56.37  & 42.10       \\
up-level0-repeat1-vit-block1-out     & 57.38       & 42.25       & 57.20  & 43.10       \\
up-level0-repeat1-vit-block3-cross-q & 56.55       & 40.50       & 55.00  & 42.50       \\
up-level0-repeat1-vit-block3-out     & 56.41       & 40.97       & 57.54  & 42.92       \\
up-level0-repeat1-vit-block5-cross-q & 54.92       & 39.66       & 54.89  & 40.79       \\
up-level0-repeat1-vit-block5-out     & 55.86       & 40.66       & 56.32  & 42.48       \\
up-level0-repeat1-vit-block7-cross-q & 51.63       & 39.01       & 52.72  & 38.37       \\
up-level0-repeat1-vit-block7-out     & 53.97       & 39.79       & 55.90  & 41.34       \\
up-level0-repeat1-vit-block9-cross-q & 50.27       & 36.42       & 36.09  & 24.81       \\
up-level0-repeat1-vit-block9-out     & 53.31       & 38.58       & 53.19  & 40.30       \\
up-level0-repeat1-vit-out            & 57.89       & 38.91       & 57.00  & 42.87       \\
up-level0-repeat2-res-out            & 56.11       & 36.37       & 55.95  & 40.62       \\
up-level0-repeat2-vit-block0-cross-q & 56.09       & 36.68       & 55.56  & 41.29       \\
up-level0-repeat2-vit-block0-out     & 57.14       & 37.16       & 56.48  & 41.14       \\
up-level0-repeat2-vit-block1-cross-q & 55.38       & 34.87       & 56.15  & 39.78       \\
up-level0-repeat2-vit-block1-out     & 56.36       & 35.63       & 56.71  & 40.93       \\
up-level0-repeat2-vit-block3-cross-q & 55.26       & 34.40       & 54.98  & 38.70       \\
up-level0-repeat2-vit-block3-out     & 55.50       & 34.61       & 56.77  & 39.36       \\
up-level0-repeat2-vit-block5-cross-q & 52.70       & 33.16       & 54.70  & 37.93       \\
up-level0-repeat2-vit-block5-out     & 54.68       & 33.75       & 55.75  & 38.95       \\
up-level0-repeat2-vit-block7-cross-q & 51.91       & 31.97       & 52.92  & 35.58       \\
up-level0-repeat2-vit-block7-out     & 53.07       & 32.43       & 54.07  & 36.91       \\
up-level0-repeat2-vit-block9-cross-q & 49.22       & 29.34       & 41.07  & 18.48       \\
up-level0-repeat2-vit-block9-out     & 51.41       & 30.64       & 51.31  & 35.64       \\
up-level0-repeat2-vit-out            & 53.77       & 33.79       & 54.24  & 37.53       \\
up-level0-upsampler-out              & 53.21       & 32.85       & 53.97  & 36.30       \\ \bottomrule
\end{tabular}
\end{table}
\begin{table}[t]
\centering
\caption{Quantitative comparison results of SDXL and Playground v2. This table shows the results from the middle resolution.
Activation ID indicates the location of each activation in the diffusion U-Net.
The best results are in \textcolor{orange}{\textbf{bold}} font and the runner-up is \textblue{\underline{underlined}}.}
\label{tab:app-comparison-2}
\begin{tabular}{@{}lcccc@{}}
\toprule
\multirow{2}{*}{Activation   ID}   & \multicolumn{2}{c}{SDXL}        & \multicolumn{2}{c}{Playground v2} \\
                                   & Simple      & Complex     & Simple & Complex     \\ \midrule
up-level1-repeat0-res-out         & \textblue{\underline{47.84}}    & 30.12          & \textblue{\underline{47.49}}     & \textblue{\underline{31.63}}     \\
up-level1-repeat0-vit-block0-cross-q & \textcolor{orange}{\textbf{49.08}} & \textcolor{orange}{\textbf{31.75}} & \textcolor{orange}{\textbf{47.96}}  & \textcolor{orange}{\textbf{32.53}}  \\
up-level1-repeat0-vit-block0-out  & 46.74       & \textblue{\underline{31.14}} & 46.49  & 30.71       \\
up-level1-repeat0-vit-block1-cross-q & 44.78       & 30.56       & 44.21  & 28.58       \\
up-level1-repeat0-vit-block1-out     & 42.20       & 27.32       & 40.90  & 27.13       \\
up-level1-repeat0-vit-out            & 46.93       & 29.35       & 46.08  & 30.15       \\
up-level1-repeat1-res-out            & 41.36       & 26.81       & 41.51  & 28.09       \\
up-level1-repeat1-vit-block0-cross-q & 39.25       & 27.02       & 39.92  & 27.72       \\
up-level1-repeat1-vit-block0-out     & 38.64       & 25.68       & 39.47  & 27.60       \\
up-level1-repeat1-vit-block1-cross-q & 36.76       & 24.57       & 37.20  & 25.82       \\
up-level1-repeat1-vit-block1-out     & 34.74       & 23.20       & 34.88  & 24.32       \\
up-level1-repeat1-vit-out            & 39.16       & 25.48       & 39.87  & 26.82       \\
up-level1-repeat2-res-out            & 33.58       & 23.99       & 36.30  & 25.26       \\
up-level1-repeat2-vit-block0-self-k  & 31.63       & 22.98       & 33.97  & 23.88       \\
up-level1-repeat2-vit-block0-cross-q & 32.16       & 24.97       & 34.30  & 24.83       \\
up-level1-repeat2-vit-block0-out     & 31.31       & 23.51       & 34.81  & 25.27       \\
up-level1-repeat2-vit-block1-self-k  & 30.39       & 22.99       & 33.01  & 25.44       \\
up-level1-repeat2-vit-block1-cross-q & 27.65       & 22.61       & 32.40  & 24.01       \\
up-level1-repeat2-vit-block1-out     & 27.66       & 22.35       & 31.44  & 24.24       \\
up-level1-repeat2-vit-out            & 27.61       & 19.80       & 26.29  & 18.61       \\ \bottomrule
\end{tabular}
\end{table}
\begin{table}[t]
\centering
\caption{Quantitative comparison results of SDv1.5.
Activation ID indicates the location of each activation in the diffusion U-Net.
The best results are in \textcolor{orange}{\textbf{bold}} font and the runner-up is \textblue{\underline{underlined}}, both marked per resolution.}
\label{tab:app-comparison-3}
\begin{tabular}{@{}lcc@{}}
\toprule
\multicolumn{1}{c}{\multirow{2}{*}{Activation   ID}} & \multicolumn{2}{c}{SDv1.5}      \\
\multicolumn{1}{c}{}                                 & Simple         & Complex        \\ \midrule
up-level1-repeat0-res-out                            & 47.16          & 37.86          \\
up-level1-repeat0-vit-block0-cross-q                 & 47.59          & \textblue{\underline{41.13}}    \\
up-level1-repeat0-vit-block0-out                     & 48.59          & 39.75          \\
up-level1-repeat0-vit-out                            & 48.21          & 39.91          \\
up-level1-repeat1-res-out                            & 49.71          & 40.80          \\
up-level1-repeat1-vit-block0-cross-q                 & 49.04          & \textcolor{orange}{\textbf{42.80}} \\
up-level1-repeat1-vit-block0-out                     & 49.88          & 39.43          \\
up-level1-repeat1-vit-out                            & \textblue{\underline{50.23}}    & 40.98          \\
up-level1-repeat2-res-out                            & \textcolor{orange}{\textbf{50.61}} & 39.47          \\
up-level1-repeat2-vit-block0-cross-q                 & 50.10          & 38.80          \\
up-level1-repeat2-vit-block0-out                     & 49.09          & 37.44          \\
up-level1-repeat2-vit-out                            & 49.62          & 36.69          \\ \midrule
up-level2-repeat0-res-out                            & 52.43          & 37.98          \\
up-level2-repeat0-vit-block0-cross-q                 & 52.59          & \textblue{\underline{39.63}}    \\
up-level2-repeat0-vit-block0-out                     & 51.56          & 36.62          \\
up-level2-repeat0-vit-out                            & 52.29          & 38.78          \\
up-level2-repeat1-res-out                            & \textblue{\underline{53.30}}    & 36.65          \\
up-level2-repeat1-vit-block0-cross-q                 & \textcolor{orange}{\textbf{53.58}} & \textcolor{orange}{\textbf{39.82}} \\
up-level2-repeat1-vit-block0-out                     & 50.70          & 35.96          \\
up-level2-repeat1-vit-out                            & 50.50          & 35.59          \\
up-level2-repeat2-res-out                            & 48.47          & 33.70          \\
up-level2-repeat2-vit-block0-self-q                  & 45.44          & 32.45          \\
up-level2-repeat2-vit-block0-self-k                  & 45.69          & 32.71          \\
up-level2-repeat2-vit-block0-cross-q                 & 46.24          & 32.95          \\
up-level2-repeat2-vit-block0-out                     & 45.33          & 32.23          \\
up-level2-repeat2-vit-out                            & 45.04          & 28.87          \\
up-level2-upsampler-out                              & 45.07          & 29.57          \\ \midrule
up-level3-repeat0-vit-block0-self-q                  & \textcolor{orange}{\textbf{38.81}} & \textblue{\underline{25.72}}    \\
up-level3-repeat0-vit-block0-self-k                  & \textblue{\underline{37.78}}    & \textcolor{orange}{\textbf{26.04}} \\
up-level3-repeat1-vit-block0-self-q                  & 31.64          & 24.02          \\
up-level3-repeat1-vit-block0-self-k                  & 32.07          & 24.00          \\
up-level3-repeat2-vit-block0-self-q                  & 29.24          & 21.27          \\
up-level3-repeat2-vit-block0-self-k                  & 29.19          & 21.39          \\ \bottomrule
\end{tabular}
\end{table}

In this part, we present all the quantitative comparison results obtained following the protocol described in \Cref{sec:method}. 
These results are from a dataset of simple scenes (Horse-21~\cite{baranchuk2021label}) and a dataset of complex scenes (Bedroom-28~\cite{baranchuk2021label}).
Since SDXL and Playground v2 share the same U-Net architecture, their results are shown in the same tables.
We display the results from the lowest resolution in \Cref{tab:app-comparison-1} and the results from the middle resolution in \Cref{tab:app-comparison-2}.
We have also done a quantitative comparison on SDv1.5, and the results are shown in \Cref{tab:app-comparison-3}.

\clearpage


\section{Other Properties of Diffusion U-Nets}
\label{sec:app-behavior}

\subsection{High-Frequency Noises in Increment Activations}
\begin{figure}[t]
  \centering
  \includegraphics[width=.7\linewidth]{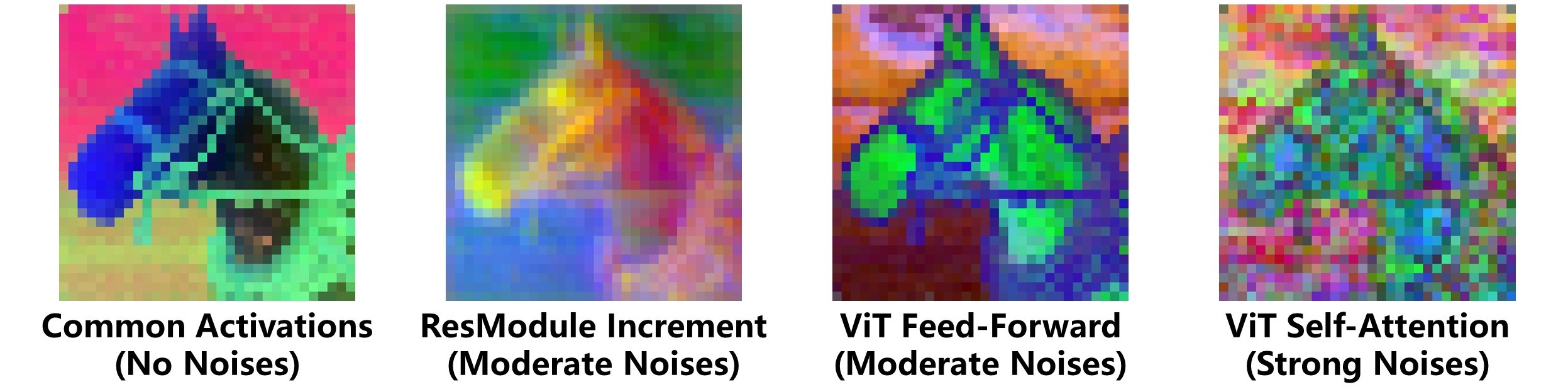}
  \caption{This visualization compares the high-frequency noises in various activations, showing three types of increment activations with strong noises.}
  \label{fig:app-mech-increment}
\end{figure}

Typically, high-frequency signals are usually noisy~\cite{carter2019activation, olah2017feature}.
This applies to the diffusion U-Net, and we can further examine what activations are more vulnerable to such noises.
To be specific, the diffusion U-Net contains many residual connection structures, and their increment activations are high-frequency signals prone to noises.
Such increment activations include:
\begin{enumerate}[label=(\roman*), leftmargin=25pt]
    \item The increment branch of ResModule. These activations are moderately noisy and thus usually less effective than inter-module activations.
    \item The feed-forward layer activations within ViTs. These activations are also moderately noisy.
    However, this results from their significantly more channels compared to other activations, which can reduce their noise magnitude.
    \item The self-attention value activations within ViTs. 
    These activations are severely noisy and suffer significant degradation as they are the nested inner increments within embedded ViTs.
\end{enumerate}
These activations are visualized in \Cref{fig:app-mech-increment} for a clearer illustration.
Additionally, the residual activations within ViTs are at the same time also increments to the main inter-module residual.
However, these activations are not obviously affected by high-frequency noise, possibly due to their dual role as ViT residuals.

\subsection{Detailed In-Resolution Granularity Change}
\begin{figure}[t]
  \centering
  \includegraphics[width=.85\linewidth]{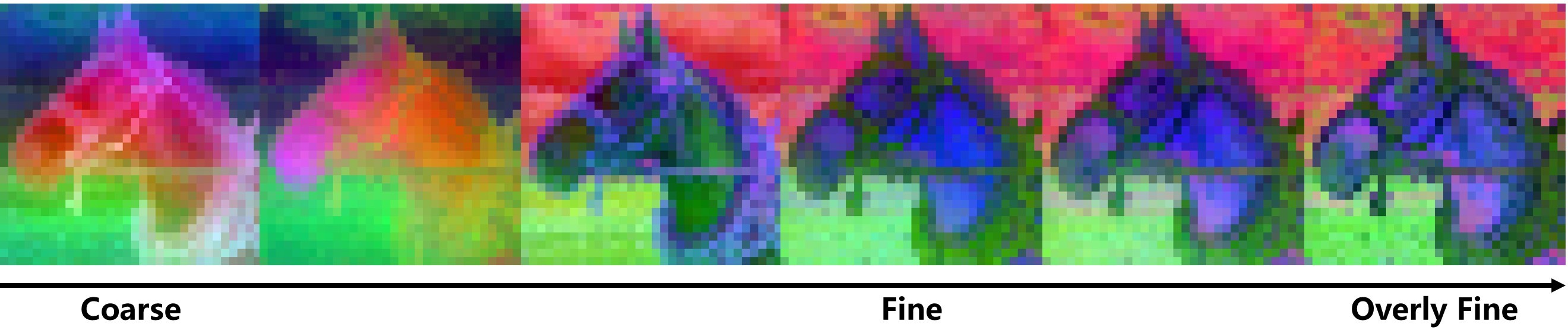}
  \caption{This visualization shows the granularity change across all inter-module activations of one resolution. At the end, the activations start to contain some slight noises, which is a sign of over-refinement.}
  \label{fig:app-mech-granularity-detail}
\end{figure}

We have previously described the existence of in-resolution granularity changes, and this section will further detail the pattern of one such change.
In one resolution of the up-stage, activations gradually shift from coarse to fine granularity, which aligns with intuition.
However, the diffusion U-Net tends to overly refine the inter-module activations, resulting in slight noises in the last few inter-module activations due to excessive detail.
This can be observed from the visualization of \Cref{fig:app-mech-granularity-detail}.
Consequently, a drop in discriminative performance is often seen near the end of one resolution.

\subsection{Collaboration between Embedded ViTs}
\begin{figure}[t]
  \centering
  \includegraphics[width=.7\linewidth]{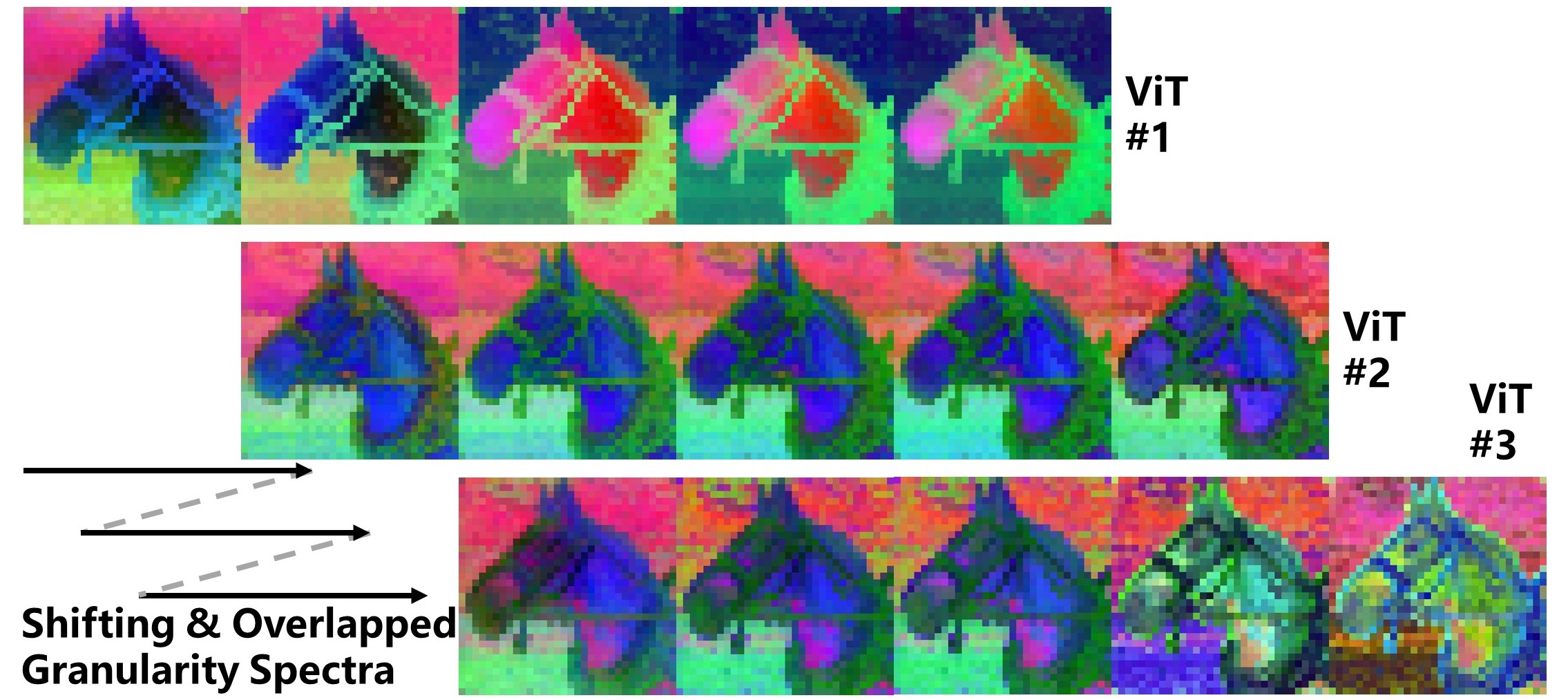}
  \caption{This visualization shows that the collaborating ViTs within one resolution contain overlapped and shifting granularity changes.}
  \label{fig:app-mech-multi-vit}
\end{figure}

Multiple ViTs exist within one resolution; for example, one resolution in the up-stage typically contains three ViTs.
These ViTs collaborate in refining the main inter-module residual.
During this process, each ViT exhibits an inner granularity change as it produces the increment activation, and the changes of collaborating ViTs form a certain pattern.
To be specific, the change of one ViT overlaps with the previous one to some extent but also shifts as a whole towards finer granularity, which is shown in \Cref{fig:app-mech-multi-vit}.

\section{Future Direction}
It might be a good future direction to focus on more challenging discrimination scenarios.
Specifically, long tail and out-of-distribution are important problems in discrimination, which can greatly hinder the performance of models that work well under i.i.d. settings~\cite{DBLP:conf/nips/WangX00CH23}.
There have been a few attempts to address this more challenging problem with diffusion models.
For example, the disentanglement property of prompts might grant diffusion features cross-domain capability~\cite{DBLP:journals/corr/abs-2307-02138}.
It is also possible to utilize diffusion models to synthesize training samples to adjust training data distribution~\cite{DBLP:conf/nips/WuZCGZHZSS23, DBLP:journals/corr/abs-2308-06739}.
Given these attempts, more efforts are still required to be put in this direction.
Moreover, there is a recent study~\cite{han2024aucseg} that might boost long tail and out-of-distribution discrimination using diffusion models.
To be specific, AUC is an evaluation metric as well as a loss function that promotes good performance on both head and tail samples, which is an important tool for long tail and out-of-distribution studies~\cite{DBLP:conf/nips/WangX00CH22, DBLP:journals/pami/WangXYHCH23}.
Conventionally, AUC is applicable to image classification but not semantic segmentation and other pixel-level tasks, but the aforementioned study manages to adopt AUC for semantic segmentation.
Given this new tool, it is now made more viable to attempt to enhance diffusion feature on long tail and out-of-distribution problems.


\section{Computation Resources}
\label{sec:app-compute}
We use Nvidia(R) RTX 3090 and Nvidia(R) RTX 4090 GPUs for the experiments, all with 24GB VRAM.
Most of our experiments, except label-scarce segmentation, require no additional storage besides the necessary space for model checkpoints and datasets.
The label-scarce segmentation task first extracts features and stores them on the disk, and then loads them for the downstream task, which takes about 4GB.

The codes are designed to be able to run on a single GPU or less than 4 GPUs, while multiple experiments can run simultaneously if more GPUs are provided.
Each experiment on the semantic correspondence task takes about 5 hours.
Each experiment on the large-scale semantic segmentation task takes 2 to 3 days, depending on the dataset.
Each experiment on the label-scarce segmentation task takes about 1 hour, but we repeat on 5 random splits following \cite{baranchuk2021label}, which increases the overall time.
All the experiments in sum can be done within two weeks.

Our quantitative comparison is based on the label-scarce segmentation task, each run taking about 40 minutes.
The time is shorter because the quantitative comparison evaluates each activation individually.
This comparison uses large storage, which is about 45GB per model per dataset.
With the help of qualitative filtering, we can finish the quantitative comparison of one model on one dataset within 2 days.

\section{Asset License}
\label{sec:app-license}
\begin{itemize}[leftmargin=10pt]
    \item SPair-71k: Available at \url{https://cvlab.postech.ac.kr/research/SPair-71k/}.
    \item Label-scarce segmentation datasets (sourced from LSUN): Available at \url{https://github.com/fyu/lsun}.
    \item ADE20K: Custom (research-only, non-commercial), at \url{https://groups.csail.mit.edu/vision/datasets/ADE20K/terms/}.
    \item CityScapes: Custom, at \url{https://www.cityscapes-dataset.com/license/}.
\end{itemize}


\newpage
\section*{NeurIPS Paper Checklist}

\begin{enumerate}

\item {\bf Claims}
    \item[] Question: Do the main claims made in the abstract and introduction accurately reflect the paper's contributions and scope?
    \item[] Answer: \answerYes{} 
    \item[] Justification: We clearly state our contributions in the abstract and introduction.

\item {\bf Limitations}
    \item[] Question: Does the paper discuss the limitations of the work performed by the authors?
    \item[] Answer: \answerYes{} 
    \item[] Justification: We discuss the limitations in \Cref{sec:conclusion}, which are mainly related to DiT models~\cite{Peebles23dit}.

\item {\bf Theory Assumptions and Proofs}
    \item[] Question: For each theoretical result, does the paper provide the full set of assumptions and a complete (and correct) proof?
    \item[] Answer: \answerNA{} 
    \item[] Justification: We do not make theoretical contributions.

    \item {\bf Experimental Result Reproducibility}
    \item[] Question: Does the paper fully disclose all the information needed to reproduce the main experimental results of the paper to the extent that it affects the main claims and/or conclusions of the paper (regardless of whether the code and data are provided or not)?
    \item[] Answer: \answerYes{} 
    \item[] Justification: The most important information to reproduce our results is the exact activations we select as features, and it is detailed in \Cref{sec:app-selection}.

\item {\bf Open access to data and code}
    \item[] Question: Does the paper provide open access to the data and code, with sufficient instructions to faithfully reproduce the main experimental results, as described in supplemental material?
    \item[] Answer: \answerNo{} 
    \item[] Justification: We will release the codes after acceptance. Besides, we have provided the necessary information for reproducing the experiments in this manuscript.

\item {\bf Experimental Setting/Details}
    \item[] Question: Does the paper specify all the training and test details (e.g., data splits, hyperparameters, how they were chosen, type of optimizer, etc.) necessary to understand the results?
    \item[] Answer: \answerYes{} 
    \item[] Justification: We have included the experimental details in \Cref{sec:app-detail}.

\item {\bf Experiment Statistical Significance}
    \item[] Question: Does the paper report error bars suitably and correctly defined or other appropriate information about the statistical significance of the experiments?
    \item[] Answer: \answerYes{} 
    \item[] Justification: We provide error bars for some experimental results. Others are omitted as it can too computationally expensive, such as about two days per run.

\item {\bf Experiments Compute Resources}
    \item[] Question: For each experiment, does the paper provide sufficient information on the computer resources (type of compute workers, memory, time of execution) needed to reproduce the experiments?
    \item[] Answer: \answerYes{} 
    \item[] Justification: We provide information in \Cref{sec:app-compute}.
    
\item {\bf Code Of Ethics}
    \item[] Question: Does the research conducted in the paper conform, in every respect, with the NeurIPS Code of Ethics \url{https://neurips.cc/public/EthicsGuidelines}?
    \item[] Answer: \answerYes{} 
    \item[] Justification: We conform with the NeurIPS Code of Ethics.

\item {\bf Broader Impacts}
    \item[] Question: Does the paper discuss both potential positive societal impacts and negative societal impacts of the work performed?
    \item[] Answer: \answerNA{} 
    \item[] Justification: There is no socetal impact.
    
\item {\bf Safeguards}
    \item[] Question: Does the paper describe safeguards that have been put in place for responsible release of data or models that have a high risk for misuse (e.g., pretrained language models, image generators, or scraped datasets)?
    \item[] Answer: \answerNA{} 
    \item[] Justification: The paper poses no such risks.

\item {\bf Licenses for existing assets}
    \item[] Question: Are the creators or original owners of assets (e.g., code, data, models), used in the paper, properly credited and are the license and terms of use explicitly mentioned and properly respected?
    \item[] Answer: \answerYes{} 
    \item[] Justification: We cite the original papers where related assets are first mentioned and list their license in \Cref{sec:app-license}.

\item {\bf New Assets}
    \item[] Question: Are new assets introduced in the paper well documented and is the documentation provided alongside the assets?
    \item[] Answer: \answerNA{} 
    \item[] Justification: We do not introduce new assets.

\item {\bf Crowdsourcing and Research with Human Subjects}
    \item[] Question: For crowdsourcing experiments and research with human subjects, does the paper include the full text of instructions given to participants and screenshots, if applicable, as well as details about compensation (if any)? 
    \item[] Answer: \answerNA{} 
    \item[] Justification: The paper does not involve crowdsourcing nor research with human subjects.

\item {\bf Institutional Review Board (IRB) Approvals or Equivalent for Research with Human Subjects}
    \item[] Question: Does the paper describe potential risks incurred by study participants, whether such risks were disclosed to the subjects, and whether Institutional Review Board (IRB) approvals (or an equivalent approval/review based on the requirements of your country or institution) were obtained?
    \item[] Answer: \answerNA{} 
    \item[] Justification: The paper does not involve human subjects.

\end{enumerate}

\end{document}